\documentclass[journal]{IEEEtran}

\usepackage{cite} 

\ifCLASSINFOpdf
\usepackage[pdftex]{graphicx}
\graphicspath{{images/}}  
\DeclareGraphicsExtensions{.pdf,.jpeg,.png}
\else 
\usepackage[dvips]{graphicx}
\graphicspath{{images/}}
\DeclareGraphicsExtensions{.eps}
\fi  

\usepackage[cmex10]{amsmath}
\usepackage{amssymb}
\usepackage{amsthm}

\usepackage{amsbsy}
\usepackage[mathcal]{eucal}

\usepackage{algorithm}
\usepackage{algpseudocode} 

\usepackage{xspace}
\algnewcommand{\algorithmicgoto}{\textbf{go to}}%
\algnewcommand{\Goto}{\algorithmicgoto\xspace}%
\algnewcommand{\Label}{\State\unskip}
\algnewcommand{\Break}{\State \textbf{break}}

\usepackage{tabularx}

\usepackage{array} 
\newcolumntype{$}{>{\global\let\currentrowstyle\relax}}
\newcolumntype{^}{>{\currentrowstyle}}
\newcommand{\rowstyle}[1]{\gdef\currentrowstyle{#1}%
  #1\ignorespaces
}

\usepackage{mathrsfs}
\DeclareMathAlphabet{\mathpzc}{OT1}{pzc}{m}{it}

\usepackage[caption=false,font=footnotesize]{subfig}

\usepackage{url}

\usepackage{fancybox}

\usepackage{calc}
\newsavebox\myboxa

\usepackage{overpic}
\usepackage{psfrag}

\usepackage{dblfloatfix}

\usepackage{txfonts}
\usepackage{textcomp}

\usepackage{stmaryrd}

\usepackage{pbox}

\usepackage{diagbox}

\usepackage[x11names]{xcolor}
\usepackage{colortbl}

\usepackage{hhline}

\usepackage{multirow}

\usepackage{makecell}

\usepackage{arydshln}

\def\arrvline{\hfil\kern\arraycolsep\vline\kern-\arraycolsep\hfilneg}

\usepackage{booktabs}

\usepackage{pifont}
\usepackage{soul}

\usepackage[export]{adjustbox}

\usepackage{enumitem}

\usepackage{bbding}

\usepackage{anyfontsize}

\usepackage[colorlinks=false]{hyperref}

\usepackage{setspace}

\makeatletter
\newsavebox\saved@arstrutbox
\newcommand*{\setarstrut}[1]{%
  \noalign{%
    \begingroup
    \global\setbox\saved@arstrutbox\copy\@arstrutbox
    #1%
    \global\setbox\@arstrutbox\hbox{%
      \vrule \@height\arraystretch\ht\strutbox
      \@depth\arraystretch \dp\strutbox
      \@width\z@
    }%
    \endgroup
  }%
}
\newcommand*{\restorearstrut}{%
  \noalign{%
    \global\setbox\@arstrutbox\copy\saved@arstrutbox
  }%
}
\makeatother

\makeatletter
\newcommand{\algmultiline}[1]{%
  \begin{tabularx}{\dimexpr\linewidth-\ALG@thistlm}[t]{@{}X@{}}
    #1
  \end{tabularx}
}
\makeatother

\makeatletter
\newcommand{\pinkbox}{%
  \fboxsep0pt
  \colorbox{LightPink1}{\begin{minipage}{1.5em}
      \textcolor{LightPink1}{C}
    \end{minipage}}
}
\makeatother
\makeatletter
\newcommand{\greenbox}{%
  \fboxsep0pt
  \colorbox{DarkSeaGreen1}{\begin{minipage}{1.5em}
      \textcolor{DarkSeaGreen1}{C}
    \end{minipage}}
}
\makeatother
\makeatletter
\newcommand{\bluebox}{%
  \fboxsep0pt
  \colorbox{LightCyan2}{\begin{minipage}{1.5em}
      \textcolor{LightCyan2}{C}
    \end{minipage}}
}
\makeatother

\makeatletter
\def\mathcolor#1#{\@mathcolor{#1}}
\def\@mathcolor#1#2#3{%
  \protect\leavevmode
  \begingroup
  \color#1{#2}#3%
  \endgroup
}
\makeatother

\DeclareFontFamily{U}{mathb}{\hyphenchar\font45}
\DeclareFontShape{U}{mathb}{m}{n}{
  <5> <6> <7> <8> <9> <10> gen * mathb
  <10.95> mathb10 <12> <14.4> <17.28> <20.74> <24.88> mathb12
}{}
\DeclareSymbolFont{mathb}{U}{mathb}{m}{n}
\DeclareFontSubstitution{U}{mathb}{m}{n}
\DeclareMathSymbol{\smalltriangleup}   {2}{mathb}{"98}
\DeclareMathSymbol{\smalltriangledown} {2}{mathb}{"99}
\DeclareMathSymbol{\blacktriangleup}   {2}{mathb}{"9C}

\newcommand{\uunderset}[2]{%
  \underset{%
    \text{\raisebox{0.6ex}{\smash{\fontsize{6}{6}$#1$}}}
  }{#2}%
}
\newcommand{\uoverset}[2]{%
  \overset{%
    \text{\raisebox{-0.5ex}{\smash{\fontsize{6}{6}$#1$}}}
  }{#2}%
}

\makeatletter
\newcommand{\raisemath}[1]{\mathpalette{\raisem@th{#1}}}
\newcommand{\raisem@th}[3]{\raisebox{#1}{$#2#3$}}
\makeatother

\makeatletter

\newcommand{\Rmnum}[1]{\expandafter\@slowromancap\romannumeral #1@}
\makeatother

\begin{document}

\title{STD-NET: Search of Image Steganalytic Deep-learning Architecture via
  Hierarchical Tensor Decomposition}

\author{ Shunquan~Tan,~\IEEEmembership{Senior Member,~IEEE,}
  Qiushi~Li, Laiyuan~Li, Bin~Li,~\IEEEmembership{Senior Member,~IEEE,}
  and~Jiwu~Huang,~\IEEEmembership{Fellow,~IEEE}%

  \thanks{All of the authors are with the Guangdong Key Laboratory of
    Intelligent Information Processing, Shenzhen Key Laboratory of
    Media Security, Guangdong Laboratory of Artificial Intelligence
    and Digital Economy (SZ), Shenzhen Institute of Artificial
    Intelligence and Robotics for Society, China. 
    S. Tan and L. Li are with College of Computer Science and
    Software Engineering, Shenzhen University.  Q. Li, B. Li and
    J. Huang are with College of Electronic and Information
    Engineering, Shenzhen University. (email:
    tansq@szu.edu.cn).}

  \thanks{This work was supported in part by the Key-Area Research and
    Development Program of Guangdong Province~(2019B010139003),
    NSFC~(61772349, U19B2022, 61872244), Guangdong Basic and Applied
    Basic Research Foundation~(2019B151502001), and Shenzhen R\&D
    Program~(JCYJ20180305124325555). This work was also supported by
    Alibaba Group through Alibaba Innovative Research (AIR) Program.}%
}
\maketitle
\begin{abstract}
  Recent studies shows that the majority of existing deep steganalysis
  models have a large amount of redundancy, which leads to a huge
  waste of storage and computing resources. The existing model
  compression method cannot flexibly compress the convolutional layer
  in residual shortcut block so that a satisfactory shrinking rate
  cannot be obtained.  In this paper, we propose STD-NET, an
  unsupervised deep-learning architecture search approach via
  hierarchical tensor decomposition for image steganalysis.  Our
  proposed strategy will not be restricted by various residual
  connections, since this strategy does not change the number of input
  and output channels of the convolution block.  We propose a
  normalized distortion threshold to evaluate the sensitivity of each
  involved convolutional layer of the base model to guide STD-NET to
  compress target network in an efficient and unsupervised approach,
  and obtain two network structures of different shapes with low
  computation cost and similar performance compared with the original
  one.  Extensive experiments have confirmed that, on one hand,
  our model can achieve comparable or even better detection
  performance in various steganalytic scenarios due to the great
  adaptivity of the obtained network architecture. On the other hand,
  the experimental results also demonstrate that our proposed strategy
  is more efficient and can remove more redundancy compared with
  previous steganalytic network compression methods.
\end{abstract}


\bstctlcite{std_bstctl}

\section{Introduction}
\label{sec:intro}

\IEEEPARstart{S}{teganalysis} aims to reveal covert communication
established via steganography. For steganalytic frameworks ``into the
wild'', low memory/computational cost, as well as lightweight
model size are just as important as high detection performance.

For steganography, digital image is the most commonly used cover
medium, and is the main battleground of the war between steganography
and steganalysis~\cite{bohme_ast_ver1_2010}.  Over the past decade,
the so-called embedding distortion minimizing
framework~\cite{filler_tifs_2010} has reigned supreme on both spatial
and frequency domain for image steganography. Most state-of-the-art
steganographic algorithms can be categorized as additive embedding
distortion minimizing schemes, including 
HILL~\cite{li_icip_2014} and MiPOD~\cite{sedighi_tifs_2016} in spatial domain,
UERD~\cite{guo_tifs_2015} in JPEG domain, as well as
UNIWARD~\cite{holub_eurasip_2014}~(including S-UNIWARD in spatial
domain, and J-UNIWARD in JPEG domain). Research on non-additive
distortion functions has made progress both in spatial
domain~\cite{denemark_ihmmsec_2015,li_tifs_2015, zhang_tcsvt_2017} and
JPEG domain~\cite{li_ihmmsec_2018, wang_tcsvt_2021_bbc, wang_tifs_2021_bbm}.
Further on, some pioneering works have been devoted to incorporating
deep-learning and reinforcement learning models into embedding
distortion minimizing
framework~\cite{tang_spl_2017,tang_tifs_2019,yang_tifs_2020,
  yedroudj_jvcir_2020,mo_tifs_2021_mcts}.

In the arm race with steganography, steganalysis has envolved from the
old-style ``rich model'' hand-crafted features
family~\cite{fridrich_tifs_2012,denemark_wifs_2014,tang_tifs_2016,tan_tifs_2017,li_tifs_2018}
equipped with an ensemble classifier~\cite{kodovsky_tifs_2012} to
deep-learning based solutions. Started from the work of Tan and
Li~\cite{tan_apsipa_2014}, deep-learning based steganalysis has
gradually overtaken ``rich model'' features
family~\cite{qian_spie_2015,pibre_ei_2016_ver2,xu_spl_2016,ye_tifs_2017,chen_ihmmsec_2017}
in the last five years.  In~\cite{ye_tifs_2017}, Ye et al. proposed a
deep steganalytic network equipped with a pre-processing layer SRM
(spatial rich model) and a new activation function called Truncated
Linear Unit (TLU), achieving significant improvement in spatial
domain.  In \cite{xu_ihmmsec_2017}, Xu proposed a 20-layer deep
residual steganalytic network. In \cite{zeng_tifs_2018}, Zeng et
al. proposed a generic hybrid deep-learning framework aiming at
large-scale JPEG image steganalysis. Afterwards in
\cite{zeng_tifs_2019}, Zeng et al. proposed WISERNet, the wider
separate-then-reunion network specifically designed for steganalysis
of true-color images. Boroumand et al. proposed a deep residual
steganalytic network SRNet~\cite{boroumand_tifs_2019}, which is the
first end-to-end steganalytic network. It becomes a popular
deep-learning solution due to its good performance in both spatial and
frequency domain steganalysis. Inspired by
Yedroudj-Net~\cite{yedroudj_icassp_2018}, Zhang et al. proposed
another deep-learning framework with distinct advantage in
spatial-domain steganalysis in \cite{zhang_tifs_2020}. In
\cite{you_tifs_2021}, You et al. proposed SiaStegNet, a siamese CNN
(Convolutional Neural Network) framework aiming at steganalysis to
images of arbitrary size. All of the above mentioned frameworks more
or less incorporate the domain knowledge behind the ``rich model''
hand-crafted features family, leading to dramatic computational cost
and storage overheads in its residual extracting.

It is well known that aimlessly scaling up deep-learning network is
not necessarily a guarantee of better
performance~\cite{tan_icml_2019}. The scalability of deep-learning
steganalyzers is by and large relevant to their application
scenarios~\cite{ruiz_icpr_2021}. Therefore, research of network
architecture search~\cite{liu_eccv_2018} and model
compression/pruning~\cite{luo_iccv_2017,liu_iclr_2019} becomes a
magnet in recent years. In \cite{tan_tifs_2021}, the authors proposed
CALPA-NET, a channel-pruning-assisted deep residual network
architecture search approach to shrink the network structure of
existing deep-learning based steganalyzers. The obtained architecture
can achieve comparative performance with less than 2\% parameters
compared to SRNet. However, CALPA-NET still has the following
deficiencies:
\begin{itemize}
\item It is a supervised data-driven solution in which labeled stego
  samples are indispensable;
\item It cannot shrink the bottom layers linked via direct shortcut
  connections, especially the noise residual extraction blocks in
  SRNet, with satisfying shrinking rates;
\item Its performance might be mildly degraded~(1\%$\sim$2\%) especially when
  aiming at J-UNIWARD steganography.
\end{itemize}

\begin{figure*}[!t]
  \centering
  \subfloat[]{
    \label{fig:tucker_decomp}
    \raisebox{-0.5\height}{
      \begin{overpic}[width=0.4\linewidth,keepaspectratio]{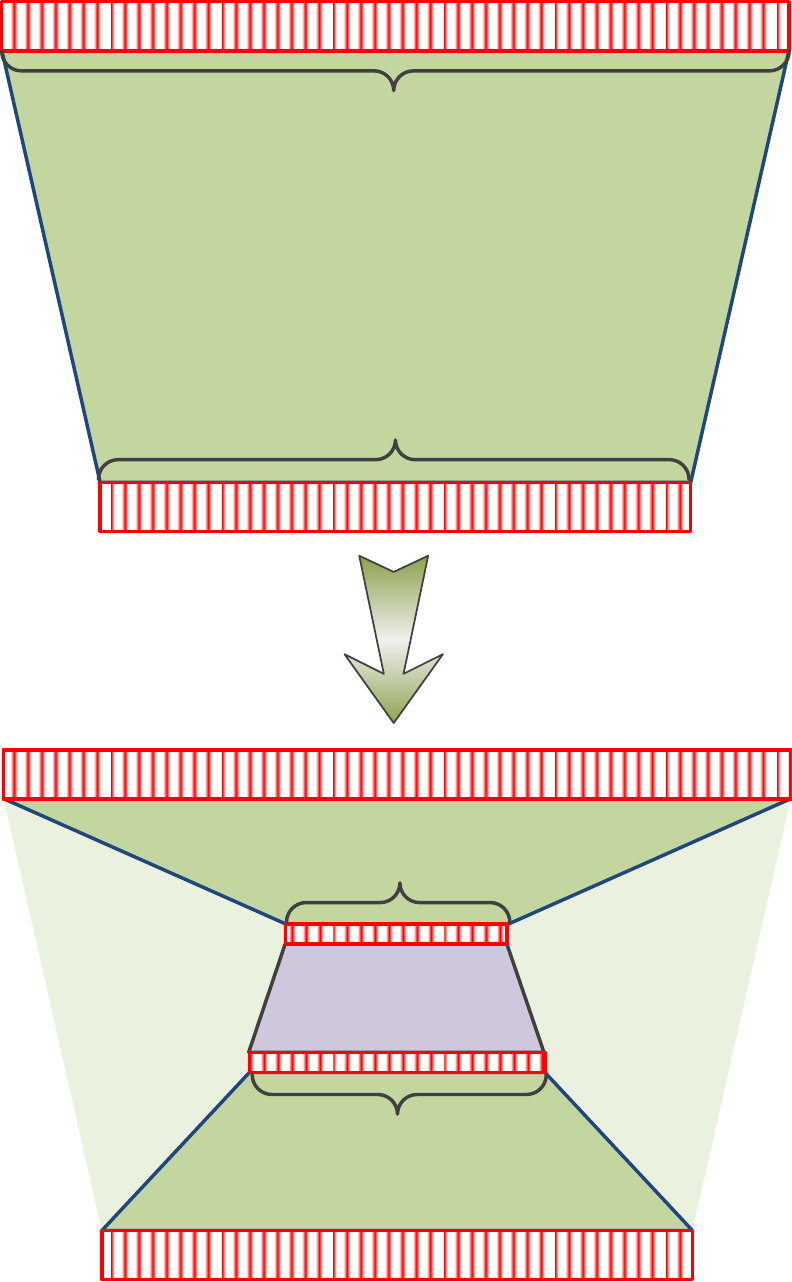}
        \put(30,89){\normalsize$J^{l}$}
        \put(30,67){\normalsize$J^{l-1}$}
        \put(30,32){\normalsize$O^{l}$} \put(30,9){\normalsize$I^{l}$}
        \put(40,77){\Huge$\boldsymbol{\mathcal{K}}^{l}$}
        \put(38,31.5){\huge$\uoverset{\raisemath{0.2ex}{\bigtriangleup}}{\boldsymbol{\mathrm{T}}}^{\raisemath{-0.6ex}{l}}$}
        \put(38,7){\huge$\uunderset{\raisemath{0.3ex}{\bigtriangleup}}{\boldsymbol{\mathrm{T}}}^{l}$}
        \put(28,20){\huge$\boldsymbol{\mathcal{G}}^{l}$}
        \put(0,74){\huge \mathcolor{red}{Layer $L_l$}}
        \put(0,8){\normalsize \parbox{13ex}{\mathcolor{red}{1st layer with $1 \times 1$ kernels}}}
        \put(6,21){\normalsize \parbox{13ex}{\mathcolor{red}{2nd layer with $D^l \times D^l$ kernels}}}
        \put(0,32){\normalsize \parbox{13ex}{\mathcolor{red}{3rd layer with $1 \times 1$ kernels}}}
      \end{overpic}  
    }
  }\hfill
  \subfloat[]{
    \label{fig:srnet_struct}
    \raisebox{-0.5\height}{\includegraphics[width=0.55\linewidth,keepaspectratio]{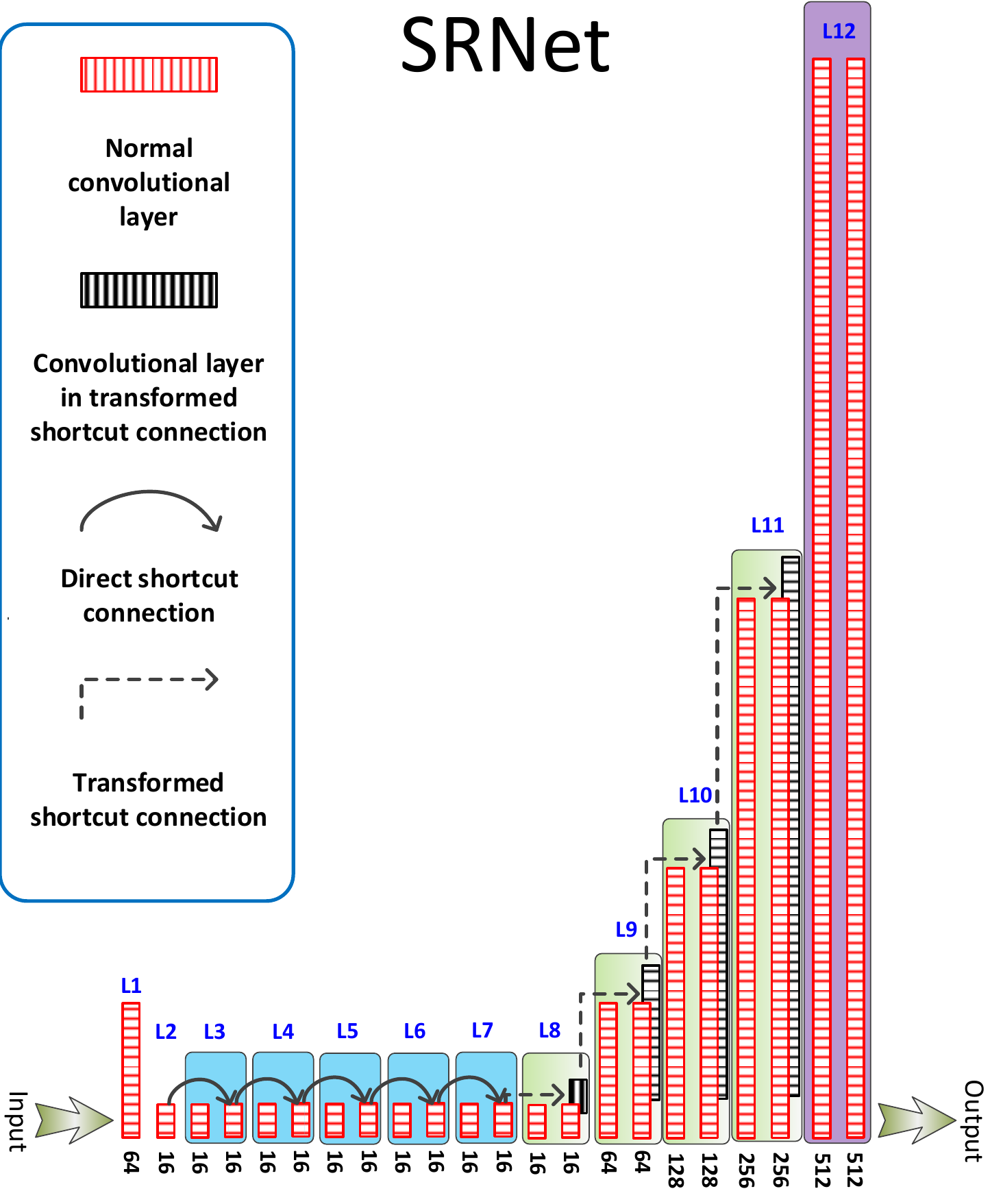}}
  }
  \caption[]{\subref{fig:tucker_decomp} The conceptual map of Tucker
    decomposition of convolution operation. \subref{fig:srnet_struct} The conceptual structure of SRNet.}
  \label{fig:tucker_decomp_srnet_struct}  
\end{figure*}
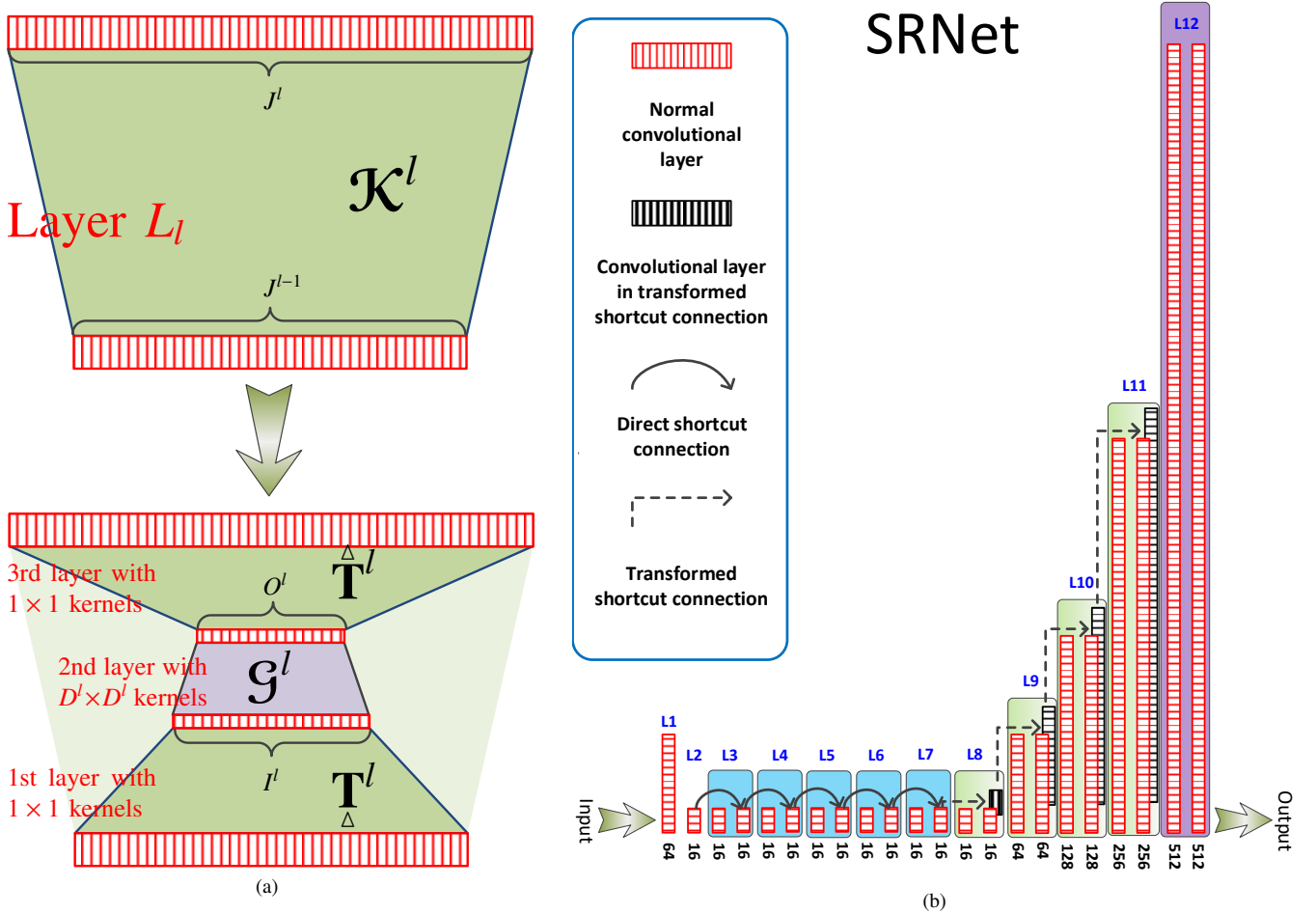 

Many researchers have pointed out that tensor analysis and
consequently hierarchical tensor decomposition can be used to dissect
and optimize CNNs, the cornerstone of deep-learning
techniques~\cite{kolda_siam_2009,cohen_arxiv_1705.02302,kim_iclr_2016,glasser_nips_2019}.
Specifically, tensor decomposition based techniques such as CP
decomposition~\cite{cohen_arxiv_1705.02302} and Tucker
decomposition~\cite{kim_iclr_2016}, which replace convolutional layers
with their low-rank matrices approximations to achieve the purpose of
effectively removing redundant information, has become a significant
branch of model compression in deep learning.  In
this paper, the authors move a step further and propose STD-NET, an
image steganalytic deep-learning architecture search approach via
hierarchical tensor decomposition.  In order to prevent the resulting
framework from being affected by specific steganography algorithm, and
to make our decomposition strategy more general, we proposed a
normalized distortion threshold to evaluate the sensitivity of each
involved convolutional layer of the target model and to guide STD-NET
to compress it in an unsupervised way.  Therefore with the help of
hierarchical tensor decomposition, starting from SRNet, we obtain a
novel image steganalytic deep-learning architecture via shrinking as
well as deepening the structure of SRNet. Compared with prior works,
STD-NET has the following novel improvements:
\begin{itemize}
\item It is an unsupervised data-driven solution in which only a few
  unlabeled samples are required, and hence a unified criterion can be applied to compress all involved convolutional layers;
\item It can significantly ``shrink'' the noise residual extraction blocks in
  SRNet and consequently dramatically save the parameters and FLOPs.
\item The extensive experiments conducted on large-scale public
  datasets have shown that STD-NET has shown fairly mild degradation
  in detection performance. Under certain scenarios its performance
  has even outperformed the original SRNet model.
\end{itemize}

The rest of the paper is organized as
follows. Sect.~\ref{sec:proposed} firstly gives a brief overview of
convolution operation, hierarchical tensor decomposition of
convolution operation as well as SRNet, the representative deep
steganalytic model. Then STD-NET, our proposed image steganalytic
deep-learning architecture search approach is described in
detail. Results of experiments conducted on large-scale public
datasets and corresponding discussions are presented in
Sect.~\ref{sec:exp}. Finally, we make a conclusion in
Sect.~\ref{sec:conclude}.
\section{Our proposed STD-NET}
\label{sec:proposed}
Throughout the text, boldface capital letters, e.g.,
$\boldsymbol{\mathrm{X}}$, denotes matrices, and boldface Euler script
letters, e.g., $\boldsymbol{\mathcal{X}}$, denotes high-order
tensors. Given a tensor $\boldsymbol{\mathcal{X}}$, its Euclidean norm
is denoted as $\lVert \boldsymbol{\mathcal{X}} \rVert$.
\subsection{Preliminaries}
\label{sec:pre}

CNN is not only the cornerstone of deep-learning techniques, but also
the infrastructure of existing deep-learning based steganalyzers. As
the name implies, the most important components of a CNN is its
convolutional layers. They take up the overwhelming majority of
learnable parameters.

Given a convolutional layer $L_l$, its convolution operation takes an
input tensor
$\boldsymbol{\mathcal{I}}^{l-1} \in \mathbb{R}^{J^{l-1} \times H^{l-1}
  \times W^{l-1}}$ which contains $J^{l-1}$ input channels with height
$H^{l-1}$ and width $W^{l-1}$, and maps it into an output tensor
$\boldsymbol{\mathcal{O}}^{l} \in \mathbb{R}^{J^{l} \times H^{l}
  \times W^{l}}$ which contains $J^{l}$ output channels with height
$H^{l}$ and width $W^{l}$. Let
$\boldsymbol{\mathcal{K}}^{l} \in \mathbb{R}^{J^{l-1} \times J^{l}
  \times D^{l} \times D^{l}}$ denote the kernel tensor.  Elementwise:
\begin{multline}
  \label{eq:conv}
  \boldsymbol{\mathcal{O}}_{i,h,w}^{l}=\sum_{j=1}^{J^{l-1}}\sum_{k=1}^{D^l}\sum_{s=1}^{D^l}\boldsymbol{\mathcal{K}}_{j,i,k,s}^{l} \cdot \boldsymbol{\mathcal{I}}_{j,h_k,w_s}^{l-1}\ ,\\
  h_k=(h-1)\cdot \Delta+k-\Psi,\ w_s=(w-1)\cdot \Delta+s-\Psi
\end{multline}
in which $\boldsymbol{\mathcal{O}}_{i,h,w}^{l}$,
$\boldsymbol{\mathcal{K}}_{j,i,k,s}^{l}$, and
$\boldsymbol{\mathcal{I}}_{j,h_k,w_s}^{l-1}$ denotes the
$(i,h,w)$\textit{-th}, the $(j,i,k,s)$\textit{-th}, and the
$(j,h_k,w_s)$\textit{-th} element of tensor
$\boldsymbol{\mathcal{O}}^{l}$, $\boldsymbol{\mathcal{K}}^{l}$, and
$\boldsymbol{\mathcal{I}}^{l-1}$ respectively. $\Delta$ is stride
and $\Psi$ is zero-padding length. In existing deep-learning based
steganalyzers, the kernel filters are usually quite small, typically
$3 \times 3$ or $5 \times 5$.

In the terminology of tensor analysis, \textit{mode} is an alias of
dimension. Given a tensor
$\boldsymbol{\mathcal{X}} \in \mathbb{R}^{I_1 \times I_2 \times \cdots
  \times I_N}$ and a matrix
$\boldsymbol{\mathrm{U}} \in \mathbb{R}^{J \times I_n}$, the
\textit{n-mode product} of $\boldsymbol{\mathcal{X}}$ and
$\boldsymbol{\mathrm{U}}$ is actually multiplying
$\boldsymbol{\mathcal{X}}$ by $\boldsymbol{\mathrm{U}}$ along its
$n$-th dimension, and is denoted as
$\boldsymbol{\mathcal{X}} \times_n \boldsymbol{\mathrm{U}} \in
\mathbb{R}^{I_1 \times \cdots \times I_{n-1} \times J \times I_{n+1}
  \times \cdots \times I_N}$. Elementwise:
\begin{equation*}
  (\boldsymbol{\mathcal{X}} \times_n
  \boldsymbol{\mathrm{U}})_{i_1\cdots i_{n-1}ji_{n+1}\cdots
    i_N}=\sum_{i_n=1}^{I_n}\boldsymbol{\mathcal{X}}_{i_1 \cdots i_n \cdots i_N}\cdot\boldsymbol{\mathrm{U}}_{ji_n}
\end{equation*}
\subsubsection{Tensor decomposition of convolution operation}
\label{sec:pre-tucker-decomp}
Here a famed variant of tensor decomposition, Tucker decomposition
which is a form of higher-order PCA~(Principal Component Analysis) is
adopted~\cite{kolda_siam_2009}.  The Tucker decomposition of
convolution operation of a given kernel tensor
$\boldsymbol{\mathcal{K}}^{l}$ can be defined as:
\begin{equation}
  \boldsymbol{\mathcal{K}}^{l}\approx
  \boldsymbol{\mathcal{G}}^{l}\times_1\uunderset{\smalltriangleup}{\boldsymbol{\mathrm{T}}}^{l}\times_2\uoverset{\smalltriangleup}{\boldsymbol{\mathrm{T}}}^{\raisemath{-0.6ex}{l}}
  \label{eq:tucker}
\end{equation}
in which the so-called \textit{core tensor}
$\boldsymbol{\mathcal{G}}^{l} \in \mathbb{R}^{I^{l} \times O^{l}
  \times D^{l} \times D^{l}}$,
$\uunderset{\smalltriangleup}{\boldsymbol{\mathrm{T}}}^{l} \in \mathbb{R}^{J^{l-1} \times I^{l}}$,
$\uoverset{\smalltriangleup}{\boldsymbol{\mathrm{T}}}^{\raisemath{-0.6ex}{l}} \in \mathbb{R}^{O^{l} \times J^{l}}$.
$\uunderset{\smalltriangleup}{\boldsymbol{\mathrm{T}}}^{l}$ and $\uoverset{\smalltriangleup}{\boldsymbol{\mathrm{T}}}^{\raisemath{-0.6ex}{l}}$ are in
orthogonal form after the
decomposition. $\boldsymbol{\mathcal{G}}^{l}$ can be regarded as a
compressed version of $\boldsymbol{\mathcal{K}}^{l}$ as long as
$I^{l}<J^{l-1}$ and $O^{l}<J^{l}$. The mode 3 and 4 are not involved
in Tucker decomposition due to the fact that they correspond to
the size of the kernel filters and are quite small, and consequently
their decomposition cannot lead in prominent cut down on computation
costs.

Please note that a \textit{n-mode product} in Eq.~\eqref{eq:tucker} is
equivalent to introducing a convolutional layer with $1 \times 1$
kernels. Therefore as illustrated in
Fig.~\ref{fig:tucker_decomp_srnet_struct}\subref{fig:tucker_decomp},
after Tucker decomposition, the convolutional layer $L_l$ with kernel
tensor $\boldsymbol{\mathcal{K}}^{l}$ is replaced by three cascading
layers:
\begin{itemize}
\item The first layer with $1 \times 1$ kernels which reduces the
  number of feature maps by passing from an input tensor of
  $\mathbb{R}^{J^{l-1} \times H^{l-1} \times W^{l-1}}$ to an output
  tensor of $\mathbb{R}^{I^{l} \times H^{l-1} \times W^{l-1}}$ since
  $I^{l}<J^{l-1}$;
\item The second layer with $D^l \times D^l$ kernels representing the
  convolution by the \textit{core tensor}
  $\boldsymbol{\mathcal{G}}^{l}$, in which the feature map tensor goes
  from the size of $\mathbb{R}^{I^{l} \times H^{l-1} \times W^{l-1}}$
  to the size of $\mathbb{R}^{O^{l} \times H^{l} \times W^{l}}$.
\item The third layer again with $1 \times 1$ kernels which
  re-increases the number of feature maps to the final output tensor
  of $\mathbb{R}^{J^{l} \times H^{l} \times W^{l}}$ since
  $O^{l}<J^{l}$.
\end{itemize}

For a given convolutional layer $L_l$, the number of parameters and
the number of FLOPs~(FLoating-point OPerations) are determined as
follows~(let $|\bullet|$ denotes the number of the corresponding
variable):
\begin{itemize}
\item $|$params$|$=$J^{l-1} \cdot J^{l}
  \cdot D^{l} \cdot D^{l}$;~\footnote{The bias, as well
    as a few parameters in other types of layers are omitted.}
\item $|$FLOPs$|$=$|$params$|$$\ \cdot\ H^{l}
  \cdot W^{l}$.
\end{itemize}
After Tucker decomposition as defined in \eqref{eq:tucker}, the
corresponding metrics of the decomposed $L_l$ can be calculated as:
\begin{itemize}
\item $|$params$|$=$J^{l-1}\cdot I^{l}+I^{l} \cdot O^{l}
  \cdot D^{l} \cdot D^{l}+O^{l} \cdot J^{l}$;
\item $|$FLOPs$|$=$|$params$|$$\ \cdot\ H^{l}
  \cdot W^{l}$.
\end{itemize}

\subsubsection{Interior structure of SRNet}
\label{sec:pre-structure-srnet}

SRNet~\cite{boroumand_tifs_2019} consists of three modules: the bottom
module tries to suppress the image contents and boost
SNR~(Signal-to-Noise Ratio)~\footnote{In the literature of
  steganalysis, image content is ``noise'' while stego noise is
  ``signal''.}, while the middle module aims at learning compact
representative features and the top module is a simple binary
``cover'' vs. ``stego'' classifier.

As illustrated in
Fig.~\ref{fig:tucker_decomp_srnet_struct}\subref{fig:srnet_struct},
following the notations in \cite{boroumand_tifs_2019}, ``L1'' and
``L2'', two hierarchical convolutional layers and the subsequent five
unpooled residuals blocks with direct shortcut connections~(from
``L3'' to ``L7'') make up the bottom module of SRNet.  From its ``L8''
up to ``L12'', the middle module gradually halves sizes of feature
maps~($256 \times 256$ $\rightarrow$ $128 \times 128$ $\rightarrow$
$64 \times 64$ $\rightarrow$ $32 \times 32$ $\rightarrow$
$16 \times 16$) as well as doubles and even quadruple~(for ``L8'')
numbers of output channels~(16 $\rightarrow$ 64 $\rightarrow$ 128
$\rightarrow$ 256 $\rightarrow$ 512) layer by layer. Among the four
middle blocks, ``L8'' to ``L11'' are with transformed shortcut
connections. The top module is a standard fully connected layer
followed by a softmax node. 
Every convolutional layer is directly followed by a batch normalization layer. 
The design of the middle module as well as
the top module of SRNet, like most of the state-of-the-art
deep-learning based steganalyzers, just simply followed the effective
recipes of the research field of computer vision and is therefore
pretty redundant. In our proposed CALPA-NET~\cite{tan_tifs_2021}, we
demonstrated that the middle/top module of SRNet can be
aggressively shrinked with comparative performance. On the contrary,
as the key part of SRNet, the bottom module is with compact design and
cannot be effectively shrinked even with CALPA-NET.

\subsection{Algorithm of our proposed STD-NET}
\label{sec:algorithm}

\subsubsection{The overall procedure}
\label{sec:algorithm-overall}

As mentioned in Sect.~\ref{sec:pre}, convolutional layer is the
most important component of CNN, taking up the majority of learnable
parameters, and there is a large amount of redundancy in it.
The diagram of overall procedure is shown in Fig. 2. According to CALPA-NET~\cite{tan_tifs_2021}, there is a lot of redundancy in the layers that can compact feature representation of SRNet (from ``L9'' to ``L12''), so we set the number of output channels for these layers to 64, and named the intermediate model as SRNetC64.
Firstly, the SRNetC64 model $N$ is trained using the original training protocol.
Then the well-trained $N$ is traversed from bottom to top and the
input/output channel number for the resulting core tensor of every
involved convolutional layer is determined in an unsupervised
data-driven manner (see Sect.~\ref{sec:algorithm-criterion}).  Two
brand-new model structures (cylinder/ladder-shaped structure, see
Sect.~\ref{sec:algorithm-channel-config}) are obtained by replacing
model $N$'s kernel tensor with corresponding Tucker decomposed
version. The resulting model can be trained in two different ways. The
first one is preserving the model parameters after Tucker
decomposition, and the other one is resetting all of the model
weights~(including those in all the $\boldsymbol{\mathcal{G}}^{l}$,
$\uunderset{\smalltriangleup}{\boldsymbol{\mathrm{T}}}^{l}$, and
$\uoverset{\smalltriangleup}{\boldsymbol{\mathrm{T}}}^{\raisemath{-0.6ex}{l}}$)
and then training it from scratch. Since the first layer of SRNet is
more sensitive to compression, the following experiment will not
decompose the first convolutional layer.

\begin{figure}[!t]
  \centering
  \includegraphics[width=0.6\columnwidth,keepaspectratio]{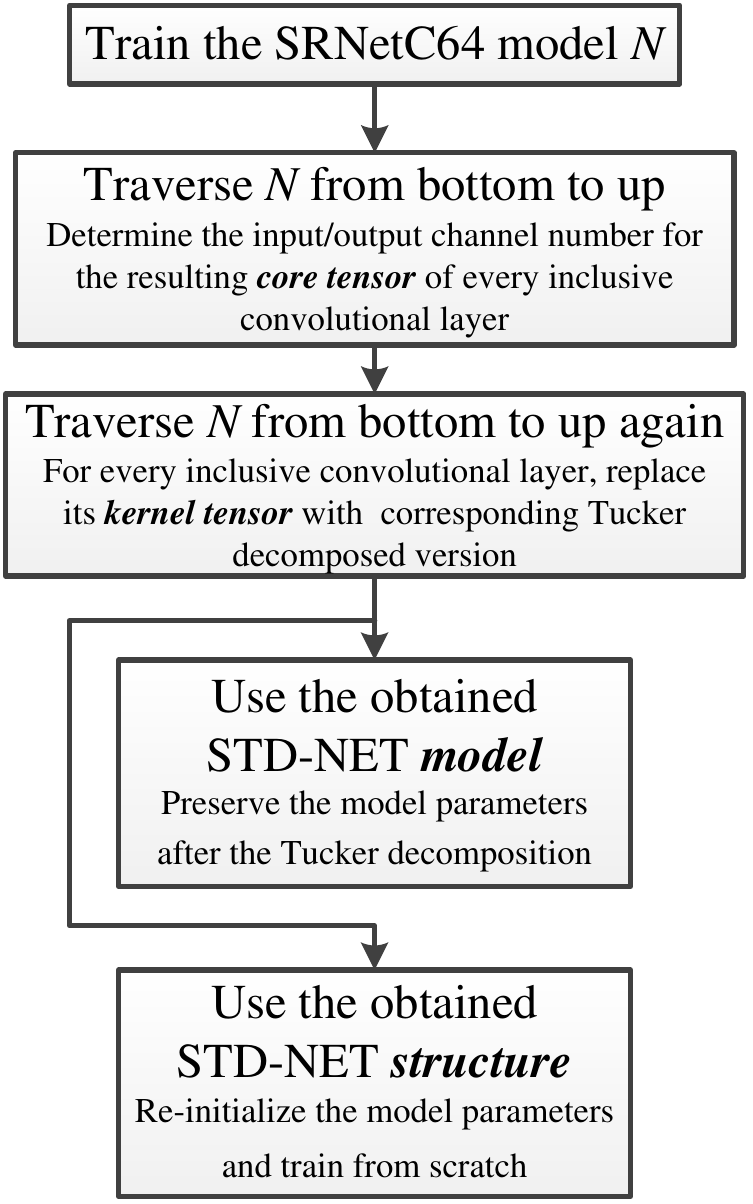} 
  \caption[]{The overall STD-NET diagram.}
  \label{fig:std_diagram}
\end{figure} 

\subsubsection{The proposed unsupervised tensor decomposition criterion}
\label{sec:algorithm-criterion}


Since a direct residual shortcut requires the number of channels at both ends to be the same, the existing channel-based network pruning strategies cannot efficiently prune both ends of a given residual shortcut with a unified criterion. On the contrary, with Tucker decomposition,
our proposed STD-NET can guarantee that 
the numbers of input and output channels of the resulting convolutional group are the same as the corresponding convolutional layer prior to the decomposition. In other words, the compression for every involved convolutional layer is independent and will not be affected by the direct residual connections in the network. All the convolutional layers can be treated equally, regardless of the residual modules they belong to.
Therefore, our proposed STD-NET approach is more concise and efficient than CALPA-NET.

It should be noted that the determination of the input/output channel number of the core tensor directly affects the final efficiency of the network compression. Therefore, we define a normalized distortion threshold:
\begin{equation}
  \mathscr{T}_d^l=\frac{\lVert \boldsymbol{\widehat{\mathcal{O}}}^{l} -
    \boldsymbol{\widehat{\mathcal{O}}}_{\boldsymbol{\mathrm{T}}}^{l}\rVert}{\lVert
    \boldsymbol{\widehat{\mathcal{O}}}^{l}\rVert}
  \label{eq:md}
\end{equation}
where $\boldsymbol{\widehat{\mathcal{O}}}^{l}$ and $\boldsymbol{\widehat{\mathcal{O}}}_{\boldsymbol{\mathrm{T}}}^{l}$ are the batch-normalized version of the features maps of the $l$-th convolutional layer in SRNetC64 model $S$ and the batch-normalized output feature maps of the corresponding Tucker decomposed convolutional group, respectively. The $\lVert \cdot \rVert$ denotes Euclidean norm.
The resulting normalized distortion threshold $\mathscr{T}_d^l$ can effectively reflect the sensitivity of each involved convolutional layer to Tucker decomposition
and assist us in choosing the appropriate number of input and output channels in an unsupervised way.



For the $l$-th convolutional layer, determining the normalized distortion $\mathscr{T}_d^l$ can be summarized as 3 steps.
Firstly, initialize the corresponding shrinking rate to 100\%.
Secondly, decrease the shrinking rate and calculate a sequence of
normalized distortion values $\boldsymbol{\mathcal{T}}^{l}$.  Finally,
choose an appropriate normalized distortion value from
$\boldsymbol{\mathcal{T}}^{l}$ and set it as $\mathscr{T}_d^l$.  The
detailed traversal algorithm for $\boldsymbol{\mathcal{T}}^{l}$ is
shown in Algorithm.~\ref{alg:distortion_threshold_determination}.
Within a certain shrinking rate change range, if the normalized
distortion values changes more drastically, it indicates that the
corresponding feature representation and even the final detection
performance will suffer greater impact.
Therefore, we need to make a trade-off between compression efficiency
and model performance so as to obtain a more lightweight model without
obvious performance drop.  Through traversing from bottom to top, we
can select an appropriate normalized distortion threshold
$\mathscr{T}_d^l$ for every involved convolutional layer $L_l$. In
Sect.~\ref{sec:ma_determine} we provide relevant experimental data of
determining the normalized distortion values. Please note that in real
scenarios, since cover images are easy to obtain, we can only use
cover images as samples for this step.

Eq.~\eqref{eq:md} actually reflects the sensitivity of each involved
convolutional layer to Tucker decomposition.  The choice of such a
distortion threshold comes directly from optimization objective of
Tucker decomposition, which minimizes changes to the output tensor
using the \textit{core tensor} with least rank.

The vanilla Tucker decomposition is implemented with manual selection
of the rank of the core tensor. Instead, as shown in
Eq.~\eqref{eq:md},we adopt an unsupervised data-driven criterion~(no
sample labels are involved in the criterion) in our proposal. The idea
comes from ThiNet~\cite{luo_iccv_2017}, one of the networking pruning
scheme utilized in our prior work
CALPA-NET~\cite{tan_tifs_2021}. ThiNet greedily prunes those channels
with smallest effect on the activation values of the next layer. With
a similar scheme, in our proposal the convolutional layer $L_l$ is
replaced by three cascading layers, and the rank of the core tensor is
tuned to guarantee that the corresponding Tucker decomposition is with
smallest effect on the activation values of the third layer with
$1 \times 1$ kernels.

\begin{algorithm}[!t]
  \caption{Normalized distortion traversal algorithm for
    $\boldsymbol{\mathcal{T}}^{l}$.}
  \label{alg:distortion_threshold_determination}
  \begin{algorithmic}[1]
    \Require A SRNetC64 model \textit{N} with the best validation
    accuracy and $\boldsymbol{\mathcal{K}}^{l}$, kernel tensor of
    the layer $L_l$ being processed, a batch of images
    $\boldsymbol{\mathcal{B}}$~(without label information) randomly
    selected from the training dataset. a pre-defined step $\epsilon$, a pre-defined lower bound of the shrinking rate $\tau$.
    
    \State Initialize $I^{'}$ and $O^{'}$, the input and output
    channel number of $\boldsymbol{\mathcal{G}}^{l}$ as:
    $I^{'}=J^{l-1}$ and $O^{'}=J^{l}$, the shrinking rate
    $\gamma=100\%$. Initialize the resulting distortion
    threshold list $\boldsymbol{\mathcal{T}}^{l}$ as an empty one-dimensional array.
    
    \State Feed \textit{N} with $\boldsymbol{\mathcal{B}}$, and then
    feedforward the input in cascaded layers bottom to up till
    $\boldsymbol{\widehat{\mathcal{O}}}^{l}$ is obtained.
    
    
    \Repeat
    
    \State Let $I^{'}=\lfloor J^{l-1}$ {\Large$\cdot$} $\gamma\rfloor$ and $O^{'}=\lfloor J^{l}$ {\Large$\cdot$} $\gamma\rfloor$.  
    
    \State $\mathscr{T}_\gamma$=\Call{CalNormDistortion}{$I^{'}$, $O^{'}$}.
    
    \State Append $\mathscr{T}_\gamma$ to $\boldsymbol{\mathcal{T}}^{l}$.
    
    
    
    \State $\gamma=\gamma-\epsilon$.
    
    \Until{$\gamma<\tau$}
    
    
    %
    %
    %
    
    \Statex
    
    \Function{CalNormDistortion}{$I$, $O$}
    \State \algmultiline{%
      Apply Tucker decomposition to $\boldsymbol{\mathcal{K}}^{l}$,
      and use
      $\boldsymbol{\mathcal{G}}^{l}\times_1\uunderset{\smalltriangleup}{\boldsymbol{\mathrm{T}}}^{l}\times_2\uoverset{\smalltriangleup}{\boldsymbol{\mathrm{T}}}^{\raisemath{-0.6ex}{l}}$
      to replace $\boldsymbol{\mathcal{K}}^{l}$ in \textit{N}. }
    
    \State \algmultiline{%
      Let \textit{N'} denotes the reconstructed
      model. Feed $\boldsymbol{\mathcal{B}}$ into
      \textit{N'} again and then feedforward it to
      get
      $\boldsymbol{\widehat{\mathcal{O}}}_{\boldsymbol{\mathrm{T}}}^{l}$. Calculate
      the normalized distortion between
      $\boldsymbol{\widehat{\mathcal{O}}}^{l}$ and
      $\boldsymbol{\widehat{\mathcal{O}}}_{\boldsymbol{\mathrm{T}}}^{l}$
      as:
      $D=\frac{\lVert
        \boldsymbol{\widehat{\mathcal{O}}}^{l} -
        \boldsymbol{\widehat{\mathcal{O}}}_{\boldsymbol{\mathrm{T}}}^{l}\rVert}{\lVert
        \boldsymbol{\widehat{\mathcal{O}}}^{l}\rVert}$.}
    
    \State \textbf{return} $D$.
    \EndFunction
  \end{algorithmic}
\end{algorithm}

\begin{algorithm}[!t]
  \caption{Input/output channel number determination algorithm for
    $\boldsymbol{\mathcal{G}}^{l}$.}
  \label{alg:input_output_channel_num_determination}
  \begin{algorithmic}[1]
    \Require A SRNetC64 model \textit{N} with the best validation
    accuracy and the kernel tensor $\boldsymbol{\mathcal{K}}^{l}$ of
    the being processed layer $L_l$ , a batch of images
    $\boldsymbol{\mathcal{B}}$~(without label information) randomly
    selected from the training dataset, a pre-determined
    distortion threshold $\mathscr{T}_d^{l}$, and a distortion margin $\varsigma$.
    
    \State Initialize $I^{'}$ and $O^{'}$, the input and output
    channel number of $\boldsymbol{\mathcal{G}}^{l}$ as:
    $I^{'}=J^{l-1}$ and $O^{'}=J^{l}$.
    
    \State \algmultiline{%
      Define the input/output channel number for a cylinder-shaped, and
      a ladder-shaped
      $\boldsymbol{\mathcal{G}}^{l}$ as
      $I_{\textrm{cylinder}}^{l}$/$O_{\textrm{cylinder}}^{l}$, and
      $I_{\textrm{ladder}}^{l}$/$O_{\textrm{ladder}}^{l}$, respectively.}
    
    \If{$J^{l-1}<J^{l}$}\label{alg:cylinder-shaped-start}
    \State $\epsilon_{I^{'}}=1$, $\epsilon_{O^{'}}=\lfloor O^{'}/I^{'} \rfloor$.
    \Else
    \State $\epsilon_{I^{'}}=\lfloor I^{'}/O^{'} \rfloor$, $\epsilon_{O^{'}}=1$.
    \EndIf
    
    \State Feed \textit{N} with $\boldsymbol{\mathcal{B}}$, and then
    feedforward the input in cascaded layers bottom to up till
    $\boldsymbol{\widehat{\mathcal{O}}}^{l}$ is obtained.
    
    \Repeat
    
    \State Let $I^{'}=I^{'}-\epsilon_{I^{'}}$ and $O^{'}=O^{'}-\epsilon_{O^{'}}$.  
    
    \State $\mathscr{T}_d^{'}$=\Call{CalNormDistortion}{$I^{'}$, $O^{'}$}.
    
    \Until{$\mathscr{T}_d^{'}>\mathscr{T}_d^{l}$}
    
    \If{$J^{l-1}\neq J^{l}$}
    \State \algmultiline{%
      The input/output channel number for a ladder-shaped
      $\boldsymbol{\mathcal{G}}^{l}$ has been determined as:
      $I_{\textrm{ladder}}^{l}=I^{‘}$ and
      $O_{\textrm{ladder}}^{l}=O^{’}$. \\
      \textbf{exit}} 
    \Else
    \State $I_{\textrm{cylinder}}^{l}=I^{‘}$ and
    $O_{\textrm{cylinder}}^{l}=O^{’}$.\label{alg:cylinder-shaped}
    \EndIf\label{alg:cylinder-shaped-end}
    
    \State Let $I^{''}=I^{'''}=I^{'}$ and
    $O^{''}=O^{'''}=O^{'}$. \label{alg:ladder-shaped-start}
    
    \Repeat
    \State Let $I^{''}=I^{''}-1$ and $O^{''}=O^{''}+1$.  
    \State $\mathscr{T}_d^{''}$=\Call{CalNormDistortion}{$I^{''}$, $O^{''}$}.
    \Until{$\mathscr{T}_d^{''}>\mathscr{T}_d^{l}+\varsigma$}
    \Repeat
    \State Let $I^{'''}=I^{'''}+1$ and $O^{'''}=O^{'''}-1$.  
    \State $\mathscr{T}_d^{'''}$=\Call{CalNormDistortion}{$I^{'''}$, $O^{'''}$}.
    \Until{$\mathscr{T}_d^{'''}>\mathscr{T}_d^{l}+\varsigma$}
    \If{$I^{''}$ {\Large$\cdot$} $O^{''} < I^{'''}$
      {\Large$\cdot$} $O^{'''}$}  
    \State $I_{\textrm{ladder}}^{l}=I^{''}$
    and $O_{\textrm{ladder}}^{l}=O^{''}$.
    \Else
    \State $I_{\textrm{ladder}}^{l}=I^{'''}$
    and $O_{\textrm{ladder}}^{l}=O^{'''}$.
    \EndIf  \label{alg:ladder-shaped-end}
    \Statex
  \end{algorithmic}
\end{algorithm}

\subsubsection{Input/output channel configuration}
\label{sec:algorithm-channel-config}

\begin{figure*}[!t]
  \centering
  \begin{overpic}[width=0.8\linewidth,keepaspectratio]{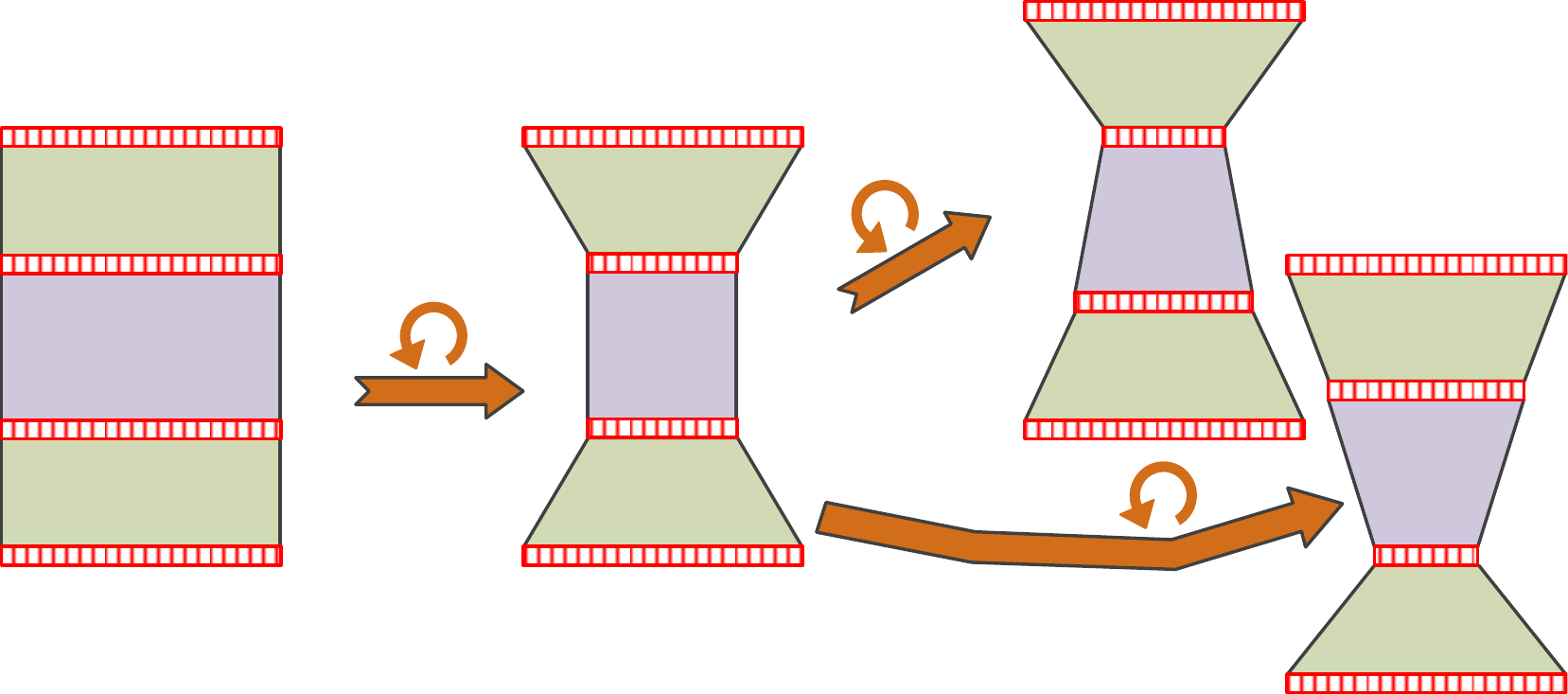}    
    \put(8,37){$J^{l}$}
    \put(4,29){$O^{'}=J^{l}$}
    \put(4,13){$I^{'}=J^{l-1}$}
    \put(6,4){$J^{l-1}$}
    \put(21,15){\footnotesize$O^{'}=O^{'}-\epsilon_{O^{'}}$}
    \put(21,12){\footnotesize$I^{'}=I^{'}-\epsilon_{I^{'}}$}
    \put(38,30){$O_{\textrm{cylinder}}^{l}$}
    \put(38,13){$I_{\textrm{cylinder}}^{l}$}
    \put(50,20){\footnotesize$I^{''}=I^{'''}=I^{'}$}
    \put(49,16){\footnotesize$O^{''}=O^{'''}=O^{'}$}
    \put(54,37){\footnotesize$O^{'''}=O^{'''}-1$}   
    \put(55,34){\footnotesize$I^{'''}=I^{'''}+1$}
    \put(70,5){\footnotesize$O^{''}=O^{''}+1$}   
    \put(70,2){\footnotesize$I^{''}=I^{''}-1$}
    \put(87,22){$O_{\textrm{ladder}}^{l}$}
    \put(88,5){$I_{\textrm{ladder}}^{l}$}
  \end{overpic}  
  \caption[]{Flow chart of input/output channel number determination
    algorithm for $\boldsymbol{\mathcal{G}}^{l}$.}
  \label{fig:alg-2-presentation}
\end{figure*}
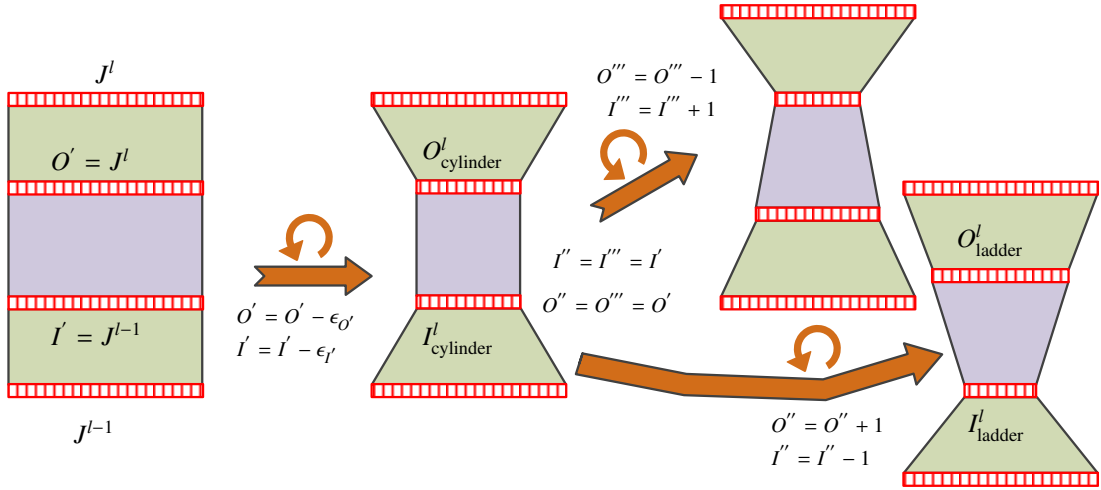

Based on the proposed unsupervised tensor decomposition criterion, we
design an algorithm of input/output channel number determination for
the core tensor $\boldsymbol{\mathcal{G}}^{l}$.

As described in
Algorithm.~\ref{alg:input_output_channel_num_determination}, for the
$l$-th convolutional layer to be decomposed in a given SRNetC64 model
$N$, the numbers of input and output channels are respectively denoted
as $J^{l-1}$ and $J^{l}$.

Firstly, the numbers of input and output channels of the core tensor
$\boldsymbol{\mathcal{G}}^{l}$ are initialized to $I'=J^{l-1}$,
$O'=J^{l}$. In this initial stage $\boldsymbol{\mathcal{G}}^{l}$ is
actually with the same size as the original kernel tensor
$\boldsymbol{\mathcal{K}}^{l}$, as illustrated in the left-hand side
of Fig.~\ref{fig:alg-2-presentation}.

Secondly, as shown in
line~\ref{alg:cylinder-shaped-start}--\ref{alg:cylinder-shaped-end} of
Algorithm~\ref{alg:input_output_channel_num_determination}, $I'$ and
$O'$ are reduced in certain steps at the same time until the
corresponding distortion threshold reaches a pre-defined normalized
distortion. Then the reduced core tensor
$\boldsymbol{\mathcal{G}}^{l}$ is obtained. In this stage,
$\epsilon_{I^{'}}$ and $\epsilon_{O^{'}}$, the reduction steps of $I'$
and $O'$ are all set to 1 for those convolutional layers in which both
input and output are with identical number of channels. $L_2$ and
$L_{9-1}$ of SRNetC64 are two exceptions since their input and output
are with different number of channels. For those exceptional layers,
$\epsilon_{I^{'}}$ and $\epsilon_{O^{'}}$ are set to make $I'$ and
$O'$ reduce in the same proportion.

Hence after this stage, for those convolutional layers in which both
input and output are with identical number of channels, the shape of
the reduced core tensor $\boldsymbol{\mathcal{G}}^{l}$ is similar to a
cylinder, as illustrated in the middle of
Fig.~\ref{fig:alg-2-presentation}. The number of input and output
channels currently searched for $\boldsymbol{\mathcal{G}}^{l}$ is
denoted as $I_{cylinder}^l$ and $O_{cylinder}^l$, respectively.

Thirdly, as shown in
line~\ref{alg:ladder-shaped-start}--\ref{alg:ladder-shaped-end} of
Algorithm.~\ref{alg:input_output_channel_num_determination}, we try to
magnify the difference between the input and output channel numbers
under the constraint that their sum is fixed, in order to further
reduce the model parameters of the obtained three cascading layers
after the Tucker decomposition. The shape of the further reduced core
tensor $\boldsymbol{\mathcal{G}}^{l}$ obtained in this way is similar
to a ladder.

In this stage two iterative cycles are conducted, in which the first
one decreases input channels and at the same time increases output
channels~($I^{''}=I^{''}-1$ and $O^{''}=O^{''}+1$), and the second one
does the opposite~($I^{'''}=I^{'''}+1$ and $O^{'''}=O^{'''}-1$). A
distortion margin $\varsigma$ is introduced in this stage. Each of the
two cycles stops once the normalized distortion exceeds the distortion
threshold plus margin, namely $\mathscr{T}_d^{l}+\varsigma$.

After this stage, as illustrated in the right-hand side of
Fig.~\ref{fig:alg-2-presentation}, two reduced core tensor are
obtained. The one with the biggest difference is selected, and the
number of its input and output channel is denoted as
$I_{\textrm{ladder}}^{l}$ and $O_{\textrm{ladder}}^{l}$, respectively.

Please note that the last stage is unnecessary for those layers whose
input and output are with different number of channels~(i.e. $L_2$ and
$L_{9-1}$ of SRNetC64). They are already with a ladder-shaped
$\boldsymbol{\mathcal{G}}^{l}$, hence the number of the input and
output channel of such a ladder-shaped $\boldsymbol{\mathcal{G}}^{l}$
is denoted as $I_{\textrm{ladder}}^{l}$ and $O_{\textrm{ladder}}^{l}$,
respectively.

After traversing the given SRNetC64 model from bottom to up, every
original involved convolutional layer from the SRNetC64 model is
replaced with Tucker-decomposed convolutional group with a determined
input/output channel number, and the corresponding STD-NET is
obtained. We name the obtained STD-NET with cylinder-shaped core
tensors~(except $L_2$ and $L_{9-1}$) the cylinder-shaped
STD-NET. Likewise, we name the one with ladder-shaped core
tensors the ladder-shaped STD-NET.

\subsubsection{Theoretical analysis regarding to ladder-shaped
  configuration}
\label{sec:analysis_ladder_shaped}
Here we prove that the ladder-shaped structure is more computational
effective than the corresponding cylinder-shaped structure.  From
Sect.~\ref{sec:pre-tucker-decomp} we have known that for a given
convolutional layer $L_l$, its parameters after Tucker
decomposition as defined in \eqref{eq:tucker} can be calculated as:
\begin{equation}
  |\textrm{params}|=J^{l-1}\cdot I^{l}+I^{l} \cdot O^{l}
  \cdot D^{l} \cdot D^{l}+O^{l} \cdot J^{l}
  \label{eq:params}
\end{equation}

If a cylinder-shaped configuration is adopted, we can get
$I^{l}=I_{\textrm{cylinder}}^{l}$ and
$O^{l}=O_{\textrm{cylinder}}^{l}$ for $\boldsymbol{\mathcal{G}}^{l}$
in line~\ref{alg:cylinder-shaped} of
Algorithm~\ref{alg:input_output_channel_num_determination}. Since
$L_2$ and $L_{15}$ have been excluded, $I^{l}=O^{l}$ and
$J^{l}=J^{l-1}$. From \eqref{eq:params} we can get:
\begin{equation}
  |\textrm{params}|=\underbrace{D^{l} \cdot
    D^{l} \cdot (I^{l} \cdot
    O^{l})}_{\text{\ding{172}}}+\underbrace{J^{l-1}\cdot (I^{l}+O^{l})}_{\text{\ding{173}}}
  \label{eq:params-2}
\end{equation}
With a fixed total channel number, namely a fixed
$\Delta=I_{\textrm{cylinder}}^{l}+O_{\textrm{cylinder}}^{l}$:
\begin{align}
  & 0 \leq (I^{l}-O^{l})^2 \nonumber\\
  \Rightarrow & 0 \leq (I^{l}+O^{l})^2-4 \cdot I^{l} \cdot O^{l} \nonumber\\
  \Rightarrow & I^{l} \cdot O^{l} \leq \frac{\Delta^2}{4}
                \label{eq:am-gm_inequality}                
\end{align}
In \eqref{eq:am-gm_inequality} the equality is hold when
$I^{l}=O^{l}$, which means that \ding{172} obtains its maximum with
$I^{l}=O^{l}$.

Since $\text{\ding{173}}=J^{l-1}\cdot \Delta$, it is fixed. Therefore
$|\textrm{params}|$ reaches its maximum with a cylinder-shaped
configuration.

Again from \eqref{eq:am-gm_inequality} we can get:
\begin{equation*}
  \frac{\Delta^2}{4}-I^{l} \cdot O^{l}=\frac{(I^{l}-O^{l})^2}{4} \ge 0  
\end{equation*}
Since $\frac{\Delta^2}{4}$ is fixed, the larger the difference between
$I^{l}$ and $O^{l}$ is, the smaller $I^{l} \cdot O^{l}$ is, and
consequently the smaller \ding{172} is. As a result, from
\eqref{eq:params-2} we can see the larger the difference between
$I^{l}$ and $O^{l}$ is, the less the parameters are required for
$L_l$. Consequently, on the basis of $I_{cylinder}^l$ and
$O_{cylinder}^l$, keeping the sum of input and output channel numbers
in the core tensor constant, magnifying the difference between them
helps to further reduce the model parameters.

\begin{figure}[!t]
  \centering
  \begin{overpic}[width=\linewidth,keepaspectratio]{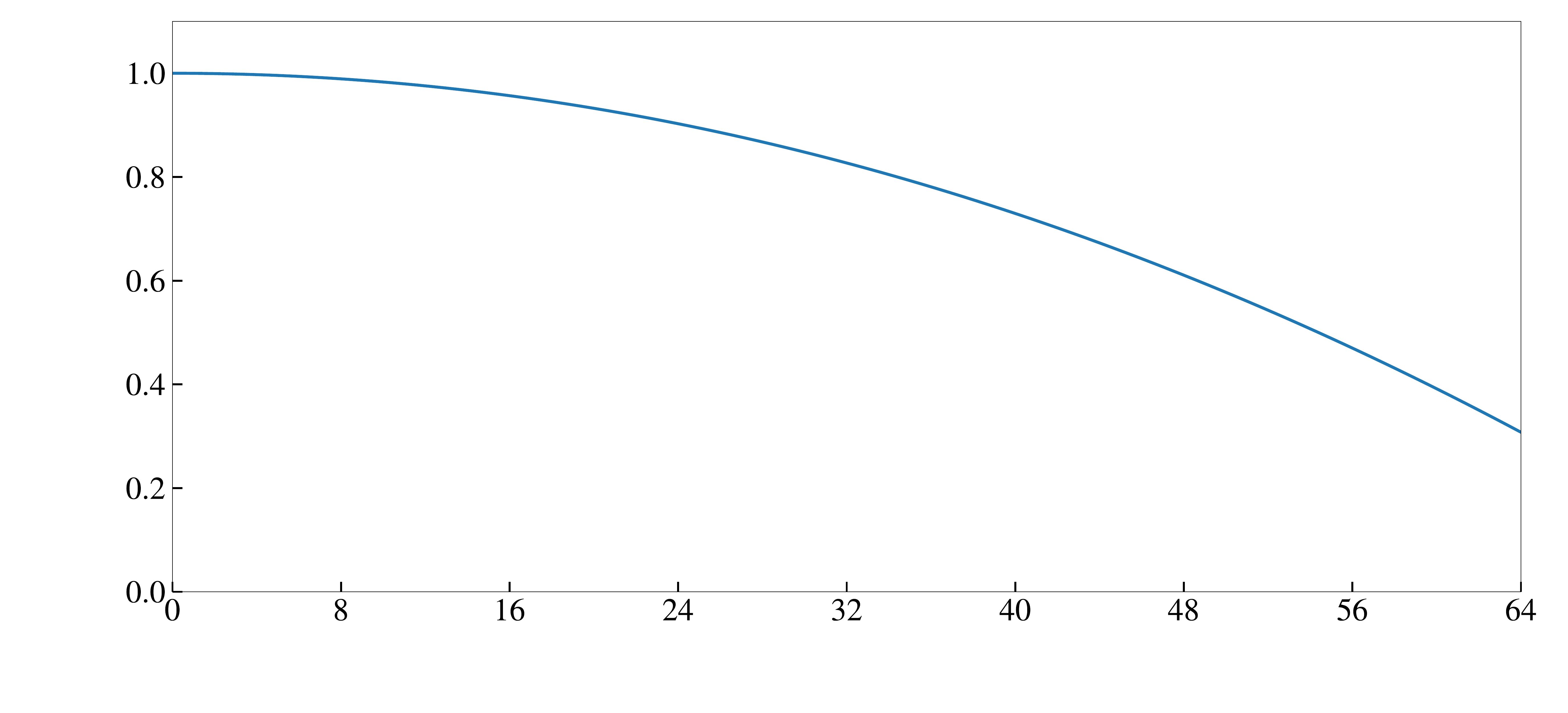}    
    \put(49,2){$\left| I^{l}-O^{l}\right|$}
    \put(2,18){\rotatebox{90}{$\frac{|\textrm{params}|_{\textrm{Ladder}}\ }{|\textrm{params}|_{\textrm{Cylinder}}}$}}
  \end{overpic}  
  \caption[]{The complexity proportion of a ladder-shaped core tensor
    to a cylinder-shaped one when the sums of their input and output
    channels are equal. }
  \label{fig:complexity_ladder_demo}
\end{figure}

Fig.~\ref{fig:complexity_ladder_demo} provides a further
demonstration. Assume that we apply Tucker decomposition to a
convolutional layer with $J^{l-1}=J^{l}=64$.  we set
$D^{l} \cdot D^{l}=9$, and the shrinking rate for searching
cylinder-shaped core tensor $\gamma=50\%$~(which means that for the
core tensor $I^{l}+O^{l}=64$). We plot the complexity proportion of
the resulting ladder-shaped core tensor to the corresponding
cylinder-shaped one when the sums of their input and output channels
are equal in Fig.~\ref{fig:complexity_ladder_demo}. From
Fig.~\ref{fig:complexity_ladder_demo} we can observe that the greater
the difference between $I^{l}$ and $O^{l}$, the less the parameters of
the ladder-shaped core tensor compared to its cylinder-shaped peer.

\subsubsection{Computational complexity analysis}
\label{sec:computational_complexity_analysis}

In order to show under what circumstances the computational complexity
after Tucker decomposition is lower than the original one, we turn to
solve the following inequality:
\begin{equation}
  \label{eq:complexity_tucker_inequality}
  J^{l-1} \cdot J^{l} \cdot D^{l} \cdot D^{l}-J^{l-1}\cdot I^{l}-I^{l}
  \cdot O^{l} \cdot D^{l} \cdot D^{l}-O^{l}  \cdot J^{l}\ge 0
\end{equation}
Since existing deep-learning based steganalyzers usually adopt fixed
small kernel filters, such as $3 \times 3$ or $5 \times 5$, in
Eq.~\eqref{eq:complexity_tucker_inequality} we treat
$D^{l} \cdot D^{l}$ as a constant. $J^{l}$ is usually an integral
multiple of $J^{l-1}$. Here for simplicity, we assume that
$J^{l-1} = J^{l}$. The proofs extended to a general case are
straightforward and are omitted for brevity. As for $I^{l}$ and
$O^{l}$, we only consider the cylinder-shaped structure, namely
$I^{l}=O^{l}$ since in Sect.~\ref{sec:analysis_ladder_shaped} we have proved
that the ladder-shaped structure is more computational effective than
the corresponding cylinder-shaped structure.

To simplify the notation we set $D^{l} \cdot D^{l}=n$,
$J^{l-1} = J^{l}=x$, and $I^{l}=O^{l}=\gamma x,\ \gamma \in [0,1]$
. So that Eq.~\eqref{eq:complexity_tucker_inequality} turns to:
\begin{equation}
  \label{eq:complexity_tucker_inequality_deduction}
  nx^2-\gamma x^2-n\gamma^2 x^2-\gamma x^2 \ge 0 \Rightarrow
  n\gamma^2+2\gamma-n \le 0
\end{equation}
Please note that now the solution of
Eq.~\eqref{eq:complexity_tucker_inequality_deduction} is actually
irrelevant to $J^{l-1}$ as well as $J^{l}$. Since
$n\gamma^2+2\gamma-n$ is a univariate quadratic function of $\gamma$
with roots $\frac{-1 \pm \sqrt{1+n^2}}{n}$,
Eq.~\eqref{eq:complexity_tucker_inequality_deduction} holds in
$\gamma \in [0, \frac{-1+\sqrt{1+n^2}}{n}]$ and its left side attains
a minimum at $\gamma=0$~(the actual minimum of the quadratic function
arrives at $-\frac{1}{n}$ which is outside the value range of
$\gamma$). Specifically, with kernel filter sizes $3 \times 3$~($n=9$)
and $5 \times 5$~($n=25$),
Eq.~\eqref{eq:complexity_tucker_inequality_deduction} holds in
$[0, 0.895]$ and $[0, 0.961]$, respectively. That is to say, for
instance, even with the small $3 \times 3$ filters used in SRNet,
Tucker decomposition brings in computational complexity reduction as
long as the shrinking rate $\gamma < 89.5\%$. Even with a conservative
shrinking rate $\gamma=50\%$, Tucker decomposition can cut the
computational complexity by $63.8\%$.

Please note that re-increasing the number of channels in STEP 3) is
just a natural consequence of Tucker decomposition. However, by doing
so, there is no longer any need to chain the modifications in the
network as in CALPA-NET and the modifications for each layer can be
made independently. Compared with the above-mentioned gain, the cost
of the introduction of STEP 3) is negligible. Please note that in
Eq.~\eqref{eq:complexity_tucker_inequality}, the part corresponding to
STEP 3) is $O^{l} \cdot J^{l}$. With the assumption and the simplified
notation used here, we can get the proportion of the part
corresponding to STEP 3) in the reduction of computational
complexity~(the left side of the inequality in
Eq.~\eqref{eq:complexity_tucker_inequality_deduction}) as
$\frac{\gamma}{n-n\gamma^2-\gamma-\gamma}$ which is next to zero.  For
instance, it gets to 0.039 with $n=9$ and a normal shrinking rate
$\gamma=30\%$.

\begin{figure*}[!htp]
  \centering
  \resizebox{0.8\linewidth}{!}{%
    \begin{minipage}[c]{\linewidth}
    \subfloat[]{
      \label{fig:ma_determine_bb}
      \begin{minipage}[c]{\linewidth}
        \begin{overpic}[width=\linewidth,keepaspectratio]{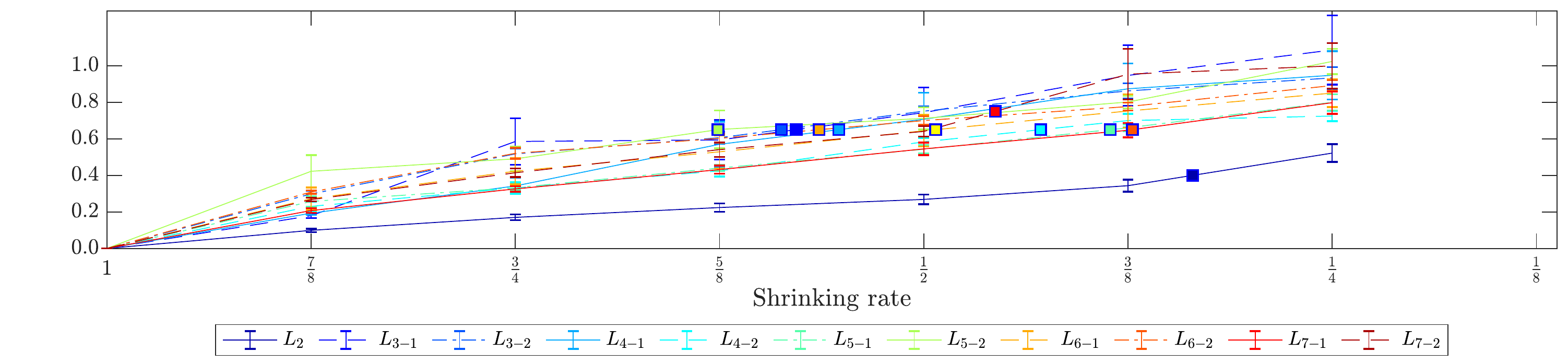}
          \put(1,12){\rotatebox{90}{$\mathscr{T}_\gamma$}}
          \put(58.8,3.8){\footnotesize$\gamma$}
        \end{overpic}  \\
        \begin{overpic}[width=\linewidth,keepaspectratio]{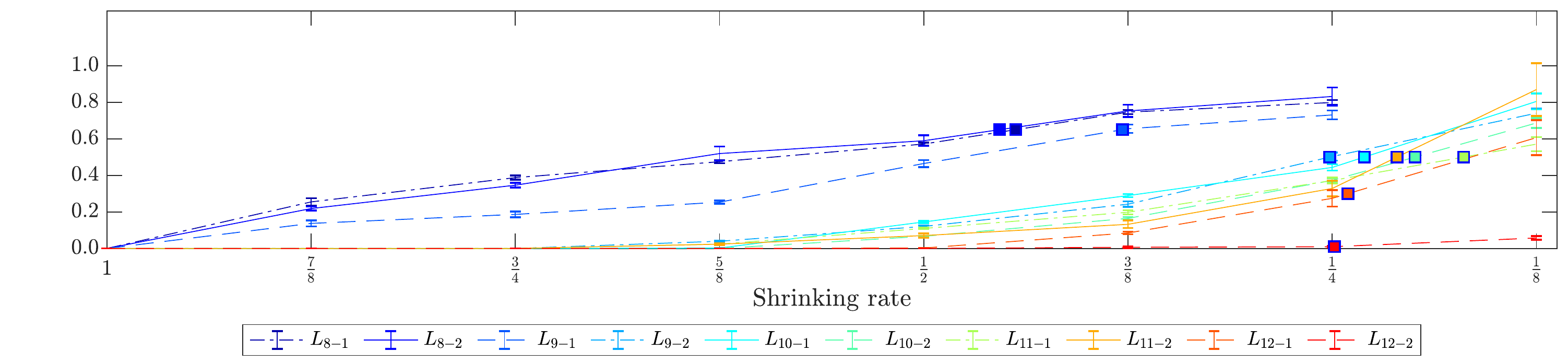}
          \put(1,12){\rotatebox{90}{$\mathscr{T}_\gamma$}}
          \put(58.8,3.8){\footnotesize$\gamma$}
        \end{overpic}
      \end{minipage}%
    }\\
    \subfloat[]{
      \label{fig:ma_determine_ala}
      \begin{minipage}[c]{\linewidth}
        \begin{overpic}[width=\linewidth,keepaspectratio]{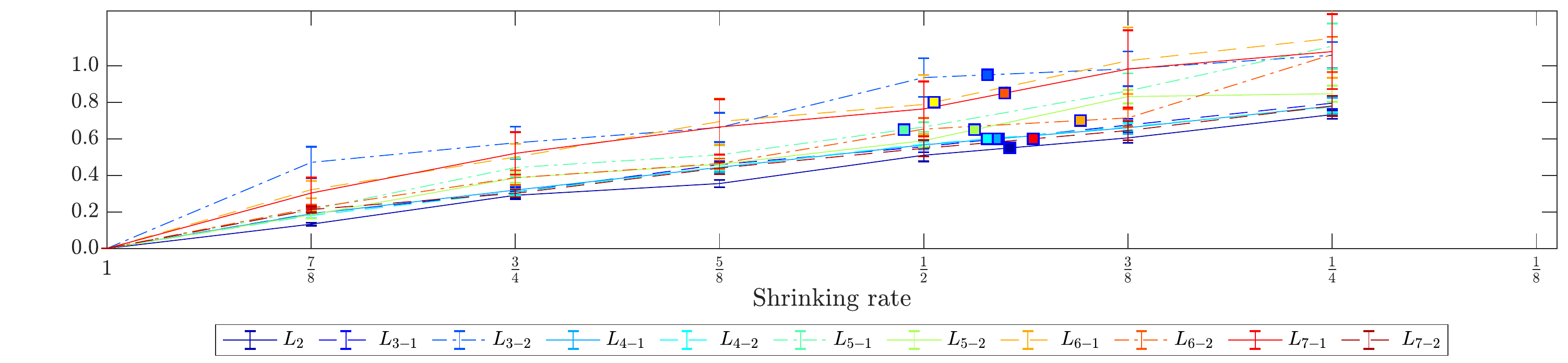}
          \put(1,12){\rotatebox{90}{$\mathscr{T}_\gamma$}}
          \put(58.8,3.8){\footnotesize$\gamma$}
        \end{overpic}  \\
        \begin{overpic}[width=\linewidth,keepaspectratio]{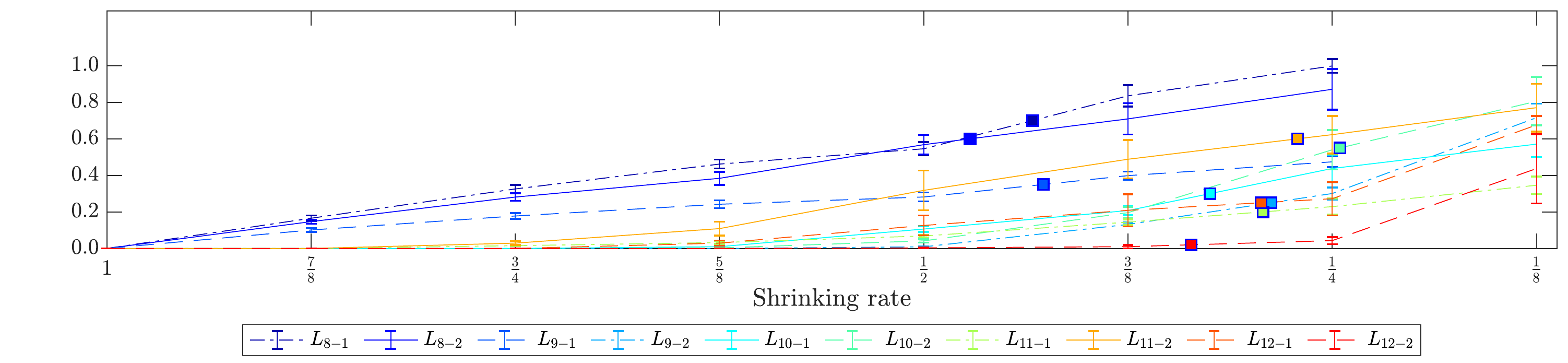}
          \put(1,12){\rotatebox{90}{$\mathscr{T}_\gamma$}}
          \put(58.8,3.8){\footnotesize$\gamma$}
        \end{overpic}
      \end{minipage}%
    }\\
  \end{minipage}%
  }
  \caption[]{\subref{fig:ma_determine_bb} and
    \subref{fig:ma_determine_ala} are the errorbars of normalized
    distortion vs. shrinking rate for every involved convolutional
    layer in the SRNetC64, conducted on JPEG-domain QF75
    BOSSBase+BOWS2 dataset and JPEG-domain QF75 ALASKA v2 dataset, 
    respectively. The square points with blue borders are the final
    determined $\mathscr{T}_d^l$. }
  \label{fig:ma_determine}
\end{figure*}
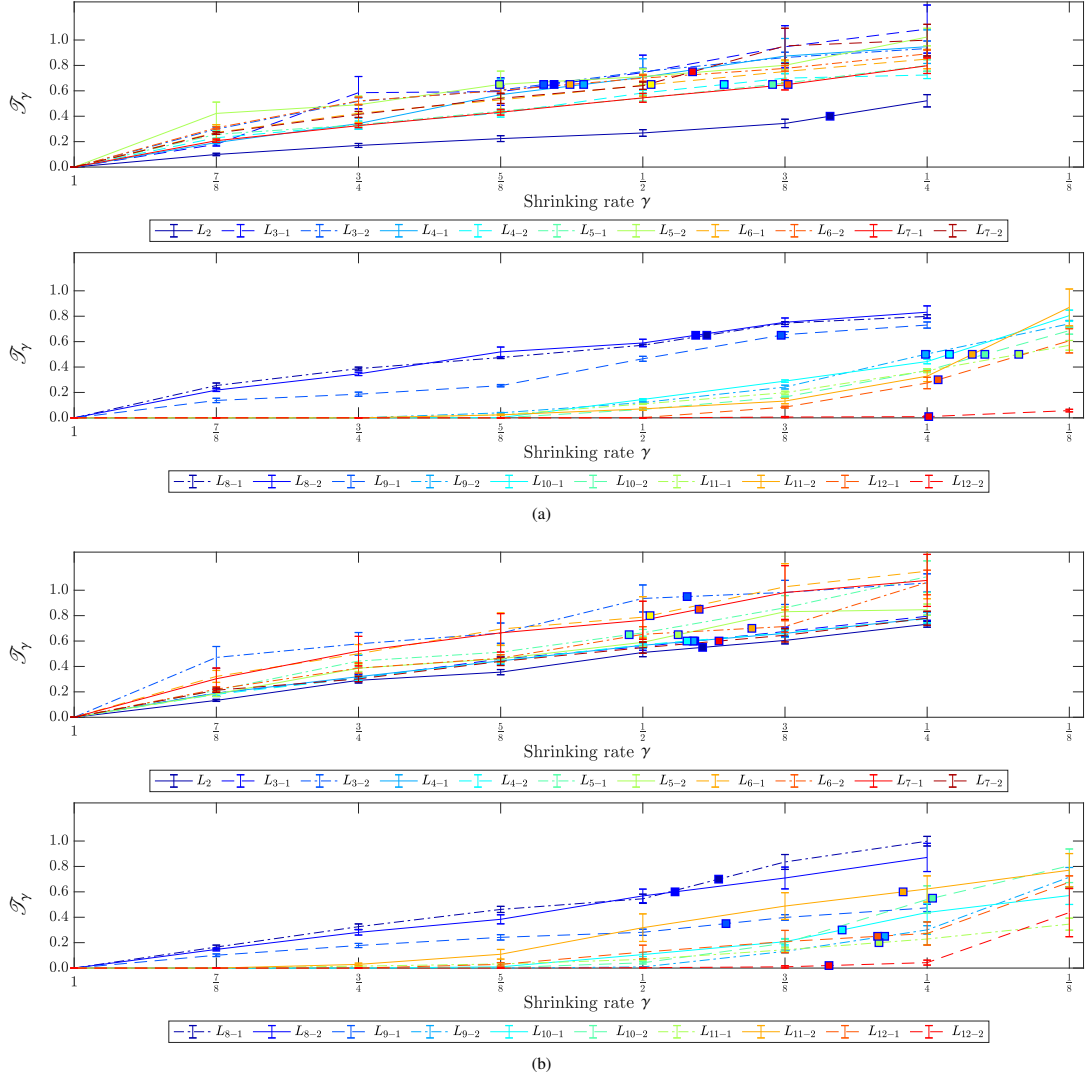

\begin{table*}[t]
  \centering
  \caption[]{Obtained structures of our proposed STD-NET under three
    different configurations and scenarios. Dimension parameters with
    \pinkbox, \greenbox and \bluebox backgrounds are from a
    cylinder-shaped~(on BOSSBase+BOWS2), a ladder-shaped~(on BOSSBase+BOWS2) and a
    cylinder-shaped configuration~(on
    ALASKAv2/QF75/256$\times$256/gray-scale dataset), respectively.}
  \adjustbox{center}{
    \resizebox{1.1\textwidth}{!}{%
      {\renewcommand{\arraystretch}{1.5}
        \setlength{\tabcolsep}{2pt}
}   
          \\        
          \hline
                                                                             & ``L1'' \arrvline &  \quad``L2'' \arrvline  & 
                                                                                                                            \multicolumn{2}{ c|}{``L3''}  &  \multicolumn{2}{ c|}{``L4''} &
                                                                                                                                                                                            \multicolumn{2}{ c|}{``L5''} & \multicolumn{2}{ c|}{``L6''} &
                                                                                                                                                                                                                                                          \multicolumn{2}{ c|}{``L7''} & \multicolumn{2}{ c|}{``L8''} &
                                                                                                                                                                                                                                                                                                                        \multicolumn{2}{ c|}{``L9''} & \multicolumn{2}{ c|}{``L10''} &
                                                                                                                                                                                                                                                                                                                                                                                       \multicolumn{2}{ c|}{``L11''} & \multicolumn{2}{ c }{``L12''} \\
          \Xhline{2\arrayrulewidth}
        \end{tabular}%
      }
    }
  }
  \label{tab:std_detailed_structure}
\end{table*}

\begin{table*}[!ht]
  \centering
  \caption[]{Comparison of parameters and FLOPs of our proposed
    STD-NET under three different configurations and scenarios. Those
    for CALPA-SRNet~(with $\varsigma=5\%$) are listed as well for
    comparison. The percentages of parameters and FLOPs of STD-NETs
    compared to CALPA-SRNet are shown in parentheses. Datas with
    \pinkbox, \greenbox and \bluebox backgrounds are from a
    cylinder-shaped~(on BOSSBase+BOWS2), a ladder-shaped~(on
    BOSSBase+BOWS2) and a cylinder-shaped configuration~(on
    ALASKAv2/QF75/256$\times$256/gray-scale dataset), respectively.}
  \adjustbox{center}{
    \resizebox{1.1\textwidth}{!}{%
      {\renewcommand{\arraystretch}{1.5}
        \setlength{\tabcolsep}{0.5pt}
        \begin{tabular}{$c^c^c^c^c^c^c^c^c^c^c^c^c^c^c^c^c^c^c^c^c^c^c}
          \Xhline{2\arrayrulewidth}
          \multicolumn{23}{ c }{\textbf{CALPA-SRNet~(with $\varsigma=5\%$)}}\\
          \hline
          \multicolumn{23}{ c }{Parameters~$(\times 10^4)$}\\
          \rowstyle{\setstretch{1}\fontsize{10}{10}\selectfont}
          6.93 & 0.05 & 0.92 & 0.18 & 0.18 & 0.14 & 0.14 & 0.05 & 0.05 & 0.14 & 0.14 & 0.20 & 0.20 & 0.14 & 0.13 & 0.37 & 0.55 & 0.11 & 0.06 & 0.41 & 0.85 & 0.56 & 1.14 \\
          \hline
          \multicolumn{23}{ c }{FLOPs~$(\times 10^8)$}\\
          \rowstyle{\setstretch{1}\fontsize{10}{10}\selectfont}
          19.70 & 0.37 & 6.03 & 1.22 & 1.22 & 0.94 & 0.94 & 0.37 & 0.37 & 0.94 & 0.94 & 1.32 & 1.32 & 0.94 & 0.88 & 0.61 & 0.90 & 0.04 & 0.02 & 0.04 & 0.08 & 0.01 & 0.02 \\
          \hline
          \multicolumn{23}{ c }{\textbf{Corresponding STD-NET structures for SRNet}}\\
          \hline
          \multicolumn{23}{ c }{Parameters~$(\times 10^4)$}\\
          \rowstyle{\setstretch{1}\fontsize{10}{10}\selectfont}
          \noindent{\begin{tabular}{m{4em}<{\centering}}
                      \cellcolor{LightPink1}{\makecell{4.97\\\textbf{(71.7\%)}}}\\
                      \cellcolor{DarkSeaGreen1}{\makecell{4.79\\\textbf{(69.0\%)}}}\\
                      \cellcolor{LightCyan2}{\makecell{8.40\\\textbf{(121.2\%)}}}\\
                    \end{tabular}} 
               &
                 \rowstyle{\setstretch{1.25}\fontsize{10}{10}\selectfont}
                 \noindent{\begin{tabular}{m{4em}<{\centering}}
                             \cellcolor{LightPink1}{\makecell{0.05}}\\
                             \cellcolor{DarkSeaGreen1}{\makecell{0.05}}\\
                             \cellcolor{LightCyan2}{\makecell{0.05}}\\
                           \end{tabular}} 
               &
                 \noindent{\begin{tabular}{m{4em}<{\centering}}
                             \cellcolor{LightPink1}{\makecell{0.44}}\\
                             \cellcolor{DarkSeaGreen1}{\makecell{0.44}}\\
                             \cellcolor{LightCyan2}{\makecell{0.29}}\\
                           \end{tabular}} 
               &
                 \noindent{\begin{tabular}{m{2.5em}<{\centering}}
                             \cellcolor{LightPink1}{\makecell{0.06}}\\
                             \cellcolor{DarkSeaGreen1}{\makecell{0.06}}\\
                             \cellcolor{LightCyan2}{\makecell{0.10}}\\
                           \end{tabular}} 
               &
                 \noindent{\begin{tabular}{m{2.5em}<{\centering}}
                             \cellcolor{LightPink1}{\makecell{0.14}}\\
                             \cellcolor{DarkSeaGreen1}{\makecell{0.12}}\\
                             \cellcolor{LightCyan2}{\makecell{0.02}}\\
                           \end{tabular}} 
               &
                 \noindent{\begin{tabular}{m{2.5em}<{\centering}}
                             \cellcolor{LightPink1}{\makecell{0.06}}\\
                             \cellcolor{DarkSeaGreen1}{\makecell{0.06}}\\
                             \cellcolor{LightCyan2}{\makecell{0.10}}\\
                           \end{tabular}} 
               &
                 \noindent{\begin{tabular}{m{2.5em}<{\centering}}
                             \cellcolor{LightPink1}{\makecell{0.08}}\\
                             \cellcolor{DarkSeaGreen1}{\makecell{0.08}}\\
                             \cellcolor{LightCyan2}{\makecell{0.16}}\\
                           \end{tabular}} 
               &
                 \noindent{\begin{tabular}{m{2.5em}<{\centering}}
                             \cellcolor{LightPink1}{\makecell{0.08}}\\
                             \cellcolor{DarkSeaGreen1}{\makecell{0.07}}\\
                             \cellcolor{LightCyan2}{\makecell{0.10}}\\
                           \end{tabular}} 
               &
                 \noindent{\begin{tabular}{m{2.5em}<{\centering}}
                             \cellcolor{LightPink1}{\makecell{0.10}}\\
                             \cellcolor{DarkSeaGreen1}{\makecell{0.09}}\\
                             \cellcolor{LightCyan2}{\makecell{0.10}}\\
                           \end{tabular}} 
               &
                 \noindent{\begin{tabular}{m{2.5em}<{\centering}}
                             \cellcolor{LightPink1}{\makecell{0.08}}\\
                             \cellcolor{DarkSeaGreen1}{\makecell{0.06}}\\
                             \cellcolor{LightCyan2}{\makecell{0.05}}\\
                           \end{tabular}} 
               &
                 \noindent{\begin{tabular}{m{2.5em}<{\centering}}
                             \cellcolor{LightPink1}{\makecell{0.14}}\\
                             \cellcolor{DarkSeaGreen1}{\makecell{0.13}}\\
                             \cellcolor{LightCyan2}{\makecell{0.10}}\\
                           \end{tabular}} 
               &
                 \noindent{\begin{tabular}{m{2.5em}<{\centering}}
                             \cellcolor{LightPink1}{\makecell{0.10}}\\
                             \cellcolor{DarkSeaGreen1}{\makecell{0.08}}\\
                             \cellcolor{LightCyan2}{\makecell{0.08}}\\
                           \end{tabular}} 
               &
                 \noindent{\begin{tabular}{m{2.5em}<{\centering}}
                             \cellcolor{LightPink1}{\makecell{0.12}}\\
                             \cellcolor{DarkSeaGreen1}{\makecell{0.10}}\\
                             \cellcolor{LightCyan2}{\makecell{0.10}}\\
                           \end{tabular}} 
               &
                 \noindent{\begin{tabular}{m{2.5em}<{\centering}}
                             \cellcolor{LightPink1}{\makecell{0.10}}\\
                             \cellcolor{DarkSeaGreen1}{\makecell{0.09}}\\
                             \cellcolor{LightCyan2}{\makecell{0.12}}\\
                           \end{tabular}} 
               &
                 \noindent{\begin{tabular}{m{2.5em}<{\centering}}
                             \cellcolor{LightPink1}{\makecell{0.05}}\\
                             \cellcolor{DarkSeaGreen1}{\makecell{0.05}}\\
                             \cellcolor{LightCyan2}{\makecell{0.10}}\\
                           \end{tabular}} 
               &
                 \noindent{\begin{tabular}{m{2.5em}<{\centering}}
                             \cellcolor{LightPink1}{\makecell{0.11}}\\
                             \cellcolor{DarkSeaGreen1}{\makecell{0.11}}\\
                             \cellcolor{LightCyan2}{\makecell{0.53}}\\
                           \end{tabular}} 
               &
                 \noindent{\begin{tabular}{m{2.5em}<{\centering}}
                             \cellcolor{LightPink1}{\makecell{0.31}}\\
                             \cellcolor{DarkSeaGreen1}{\makecell{0.31}}\\
                             \cellcolor{LightCyan2}{\makecell{1.00}}\\
                           \end{tabular}} 
               &
                 \noindent{\begin{tabular}{m{3em}<{\centering}}
                             \cellcolor{LightPink1}{\makecell{0.31}}\\
                             \cellcolor{DarkSeaGreen1}{\makecell{0.30}}\\
                             \cellcolor{LightCyan2}{\makecell{0.82}}\\
                           \end{tabular}} 
               &
                 \noindent{\begin{tabular}{m{3em}<{\centering}}
                             \cellcolor{LightPink1}{\makecell{0.24}}\\
                             \cellcolor{DarkSeaGreen1}{\makecell{0.24}}\\
                             \cellcolor{LightCyan2}{\makecell{0.47}}\\
                           \end{tabular}} 
               &
                 \noindent{\begin{tabular}{m{3em}<{\centering}}
                             \cellcolor{LightPink1}{\makecell{0.21}}\\
                             \cellcolor{DarkSeaGreen1}{\makecell{0.20}}\\
                             \cellcolor{LightCyan2}{\makecell{0.77}}\\
                           \end{tabular}} 
               &
                 \noindent{\begin{tabular}{m{3em}<{\centering}}
                             \cellcolor{LightPink1}{\makecell{0.31}}\\
                             \cellcolor{DarkSeaGreen1}{\makecell{0.31}}\\
                             \cellcolor{LightCyan2}{\makecell{0.88}}\\
                           \end{tabular}} 
               &
                 \noindent{\begin{tabular}{m{3em}<{\centering}}
                             \cellcolor{LightPink1}{\makecell{0.39}}\\
                             \cellcolor{DarkSeaGreen1}{\makecell{0.37}}\\
                             \cellcolor{LightCyan2}{\makecell{0.88}}\\
                           \end{tabular}} 
               &	
                 \noindent{\begin{tabular}{m{3em}<{\centering}}
                             \cellcolor{LightPink1}{\makecell{0.43}}\\
                             \cellcolor{DarkSeaGreen1}{\makecell{0.39}}\\
                             \cellcolor{LightCyan2}{\makecell{0.56}}\\
                           \end{tabular}} 
          \\
          \hline
          \multicolumn{23}{ c }{FLOPs~$(\times 10^8)$}\\
          \rowstyle{\setstretch{1}\fontsize{10}{10}\selectfont}
          \noindent{\begin{tabular}{m{4em}<{\centering}}
                      \cellcolor{LightPink1}{\makecell{12.00\\\textbf{(60.9\%)}}}\\
                      \cellcolor{DarkSeaGreen1}{\makecell{11.42\\\textbf{(57.9\%)}}}\\
                      \cellcolor{LightCyan2}{\makecell{13.32\\\textbf{(67.6\%)}}}\\
                    \end{tabular}} 
               &
                 \rowstyle{\setstretch{1.25}\fontsize{10}{10}\selectfont}
                 \noindent{\begin{tabular}{m{4em}<{\centering}}
                             \cellcolor{LightPink1}{\makecell{0.37}}\\
                             \cellcolor{DarkSeaGreen1}{\makecell{0.37}}\\
                             \cellcolor{LightCyan2}{\makecell{0.37}}\\
                           \end{tabular}} 
               &
                 \noindent{\begin{tabular}{m{4em}<{\centering}}
                             \cellcolor{LightPink1}{\makecell{2.93}}\\
                             \cellcolor{DarkSeaGreen1}{\makecell{2.93}}\\
                             \cellcolor{LightCyan2}{\makecell{1.91}}\\
                           \end{tabular}} 
               &
                 \noindent{\begin{tabular}{m{2.5em}<{\centering}}
                             \cellcolor{LightPink1}{\makecell{0.43}}\\
                             \cellcolor{DarkSeaGreen1}{\makecell{0.43}}\\
                             \cellcolor{LightCyan2}{\makecell{0.66}}\\
                           \end{tabular}} 
               &
                 \noindent{\begin{tabular}{m{2.5em}<{\centering}}
                             \cellcolor{LightPink1}{\makecell{0.94}}\\
                             \cellcolor{DarkSeaGreen1}{\makecell{0.85}}\\
                             \cellcolor{LightCyan2}{\makecell{0.17}}\\
                           \end{tabular}} 
               &
                 \noindent{\begin{tabular}{m{2.5em}<{\centering}}
                             \cellcolor{LightPink1}{\makecell{0.43}}\\
                             \cellcolor{DarkSeaGreen1}{\makecell{0.42}}\\
                             \cellcolor{LightCyan2}{\makecell{0.66}}\\
                           \end{tabular}} 
               &
                 \noindent{\begin{tabular}{m{2.5em}<{\centering}}
                             \cellcolor{LightPink1}{\makecell{0.54}}\\
                             \cellcolor{DarkSeaGreen1}{\makecell{0.54}}\\
                             \cellcolor{LightCyan2}{\makecell{1.10}}\\
                           \end{tabular}} 
               &
                 \noindent{\begin{tabular}{m{2.5em}<{\centering}}
                             \cellcolor{LightPink1}{\makecell{0.54}}\\
                             \cellcolor{DarkSeaGreen1}{\makecell{0.49}}\\
                             \cellcolor{LightCyan2}{\makecell{0.66}}\\
                           \end{tabular}} 
               &
                 \noindent{\begin{tabular}{m{2.5em}<{\centering}}
                             \cellcolor{LightPink1}{\makecell{0.66}}\\
                             \cellcolor{DarkSeaGreen1}{\makecell{0.64}}\\
                             \cellcolor{LightCyan2}{\makecell{0.66}}\\
                           \end{tabular}} 
               &
                 \noindent{\begin{tabular}{m{2.5em}<{\centering}}
                             \cellcolor{LightPink1}{\makecell{0.54}}\\
                             \cellcolor{DarkSeaGreen1}{\makecell{0.45}}\\
                             \cellcolor{LightCyan2}{\makecell{0.33}}\\
                           \end{tabular}} 
               &
                 \noindent{\begin{tabular}{m{2.5em}<{\centering}}
                             \cellcolor{LightPink1}{\makecell{0.94}}\\
                             \cellcolor{DarkSeaGreen1}{\makecell{0.89}}\\
                             \cellcolor{LightCyan2}{\makecell{0.66}}\\
                           \end{tabular}} 
               &
                 \noindent{\begin{tabular}{m{2.5em}<{\centering}}
                             \cellcolor{LightPink1}{\makecell{0.66}}\\
                             \cellcolor{DarkSeaGreen1}{\makecell{0.54}}\\
                             \cellcolor{LightCyan2}{\makecell{0.54}}\\
                           \end{tabular}} 
               &
                 \noindent{\begin{tabular}{m{2.5em}<{\centering}}
                             \cellcolor{LightPink1}{\makecell{0.79}}\\
                             \cellcolor{DarkSeaGreen1}{\makecell{0.70}}\\
                             \cellcolor{LightCyan2}{\makecell{0.66}}\\
                           \end{tabular}} 
               &
                 \noindent{\begin{tabular}{m{2.5em}<{\centering}}
                             \cellcolor{LightPink1}{\makecell{0.66}}\\
                             \cellcolor{DarkSeaGreen1}{\makecell{0.64}}\\
                             \cellcolor{LightCyan2}{\makecell{0.79}}\\
                           \end{tabular}} 
               &
                 \noindent{\begin{tabular}{m{2.5em}<{\centering}}
                             \cellcolor{LightPink1}{\makecell{0.33}}\\
                             \cellcolor{DarkSeaGreen1}{\makecell{0.33}}\\
                             \cellcolor{LightCyan2}{\makecell{0.66}}\\
                           \end{tabular}} 
               &
                 \noindent{\begin{tabular}{m{2.5em}<{\centering}}
                             \cellcolor{LightPink1}{\makecell{0.18}}\\
                             \cellcolor{DarkSeaGreen1}{\makecell{0.18}}\\
                             \cellcolor{LightCyan2}{\makecell{0.87}}\\
                           \end{tabular}} 
               &
                 \noindent{\begin{tabular}{m{2.5em}<{\centering}}
                             \cellcolor{LightPink1}{\makecell{0.52}}\\
                             \cellcolor{DarkSeaGreen1}{\makecell{0.51}}\\
                             \cellcolor{LightCyan2}{\makecell{1.64}}\\
                           \end{tabular}} 
               &
                 \noindent{\begin{tabular}{m{3em}<{\centering}}
                             \cellcolor{LightPink1}{\makecell{0.13}}\\
                             \cellcolor{DarkSeaGreen1}{\makecell{0.12}}\\
                             \cellcolor{LightCyan2}{\makecell{0.33}}\\
                           \end{tabular}} 
               &
                 \noindent{\begin{tabular}{m{3em}<{\centering}}
                             \cellcolor{LightPink1}{\makecell{0.10}}\\
                             \cellcolor{DarkSeaGreen1}{\makecell{0.10}}\\
                             \cellcolor{LightCyan2}{\makecell{0.19}}\\
                           \end{tabular}} 
               &
                 \noindent{\begin{tabular}{m{3em}<{\centering}}
                             \cellcolor{LightPink1}{\makecell{0.02}}\\
                             \cellcolor{DarkSeaGreen1}{\makecell{0.02}}\\
                             \cellcolor{LightCyan2}{\makecell{0.07}}\\
                           \end{tabular}} 
               &
                 \noindent{\begin{tabular}{m{3em}<{\centering}}
                             \cellcolor{LightPink1}{\makecell{0.03}}\\
                             \cellcolor{DarkSeaGreen1}{\makecell{0.03}}\\
                             \cellcolor{LightCyan2}{\makecell{0.09}}\\
                           \end{tabular}} 
               &
                 \noindent{\begin{tabular}{m{3em}<{\centering}}
                             \cellcolor{LightPink1}{\makecell{0.01}}\\
                             \cellcolor{DarkSeaGreen1}{\makecell{0.01}}\\
                             \cellcolor{LightCyan2}{\makecell{0.02}}\\
                           \end{tabular}} 
               &	
                 \noindent{\begin{tabular}{m{3em}<{\centering}}
                             \cellcolor{LightPink1}{\makecell{0.01}}\\
                             \cellcolor{DarkSeaGreen1}{\makecell{0.01}}\\
                             \cellcolor{LightCyan2}{\makecell{0.01}}\\
                           \end{tabular}} 
          \\
          \hline
               & $L_1$ & $L_2$ & $L_{3-1}$ & $L_{3-2}$ & $L_{4-1}$ & $L_{4-2}$ & $L_{5-1}$ & $L_{5-2}$ & $L_{6-1}$ & $L_{6-2}$ & $L_{7-1}$ &
                                                                                                                                             $L_{7-2}$ & $L_{8-1}$ & $L_{8-2}$ & $L_{9-1}$ & $L_{9-2}$ & $L_{10-1}$ & $L_{10-2}$ & $L_{11-1}$ & $L_{11-2}$ & $L_{12-1}$ & $L_{12-2}$\\
          \hline
          \textbf{Overall} \arrvline & \quad``L1'' \arrvline &  \quad``L2'' \arrvline  & 
                                                                                         \multicolumn{2}{ c|}{``L3''}  &  \multicolumn{2}{ c|}{``L4''} &
                                                                                                                                                         \multicolumn{2}{ c|}{``L5''} & \multicolumn{2}{ c|}{``L6''} &
                                                                                                                                                                                                                       \multicolumn{2}{ c|}{``L7''} & \multicolumn{2}{ c|}{``L8''} &
                                                                                                                                                                                                                                                                                     \multicolumn{2}{ c|}{``L9''} & \multicolumn{2}{ c|}{``L10''} &
                                                                                                                                                                                                                                                                                                                                                    \multicolumn{2}{ c|}{``L11''} & \multicolumn{2}{ c }{``L12''} \\
          \Xhline{2\arrayrulewidth}
        \end{tabular}%
      }
    }
  }
  \label{tab:std_params_flops}
\end{table*}

\section{Experiments}
\label{sec:exp}

\subsection{Experiment setup}
\label{sec:setup}

\subsubsection{Datasets}
\label{sec:data}

The primary image dataset used in our experiments is the union of BOSSBase v1.01~\cite{bas_ih_2011_boss} and BOWS2~\cite{bows2}, each of which contains 10,000 512 $\times$ 512 grayscale spatial images. All of the images were resized to 256 $\times$ 256 using Matlab function \textit{imresize}. The corresponding JPEG images were further generated with QFs (Quality Factors) 75 and 95. In our experiments, 10,000 BOWS2 images and 4,000 randomly selected BOSSBase images were used for training. Another 1,000 randomly selected BOSSBase images were for validation. The remaining 5,000 BOSSBase images were retained for testing. 

ALASKA v2~\cite{alaska2_link} is another large-scale dataset with totally 80,005 images, introduced to evaluate the performance of STD-NET. The uncompressed and JPEG compressed (with quality factors: 75 and various QF) grayscale images datasets of size 256 $\times$ 256 were downloaded from ALASKA v2's official website\footnote{https://alaska.utt.fr/\#material}. We used training and validation sets of randomly selected 56,000 and 4,000 images, respectively, and the remaining images for testing. There are no identical images between the three sets.

\subsubsection{Steganography schemes}
\label{sec:stegano}

Four representative steganographic schemes, UERD [5] and J-UNIWARD [6] for JPEG domain, and HILL [3] and S-UNIWARD [6] for spatial domain, were our attacking targets in the experiments. For JPEG steganographic algorithms, the embedding payloads were set to 0.2 and 0.4 bpnzAC (bits per non-zero AC DCT coefficient). For spatial domain steganographic algorithms, the embedding payloads were set to 0.2 and 0.4 bpp (bits per pixel).

As pointed out in \cite{sedighi_spie_2016}, the performance of non-additive schemes
varies dramatically on different cover sources. Therefore, we mainly focus on the detection of additive embedding distortion steganography algorithms. 
As for non-additive schemes, the complementary experimental results can be found in Sect.~\ref{sec:non_additive}.

\subsubsection{Detectors}
\label{sec:detector}

SRNetC64 (a variant of SRNet, see Sect.~\ref{sec:algorithm-overall})
was selected as the initial architectures of our proposed STD-NET. Our
implementation of STD-NET and its corresponding initial architectures
are based on Tensorflow~\cite{tensorflow2015-whitepaper}. Unless
otherwise specified, the initial architecture were trained with the
hyperparameters mentioned in \cite{boroumand_tifs_2019}. The batch
size in the training procedure was set to 32 (namely 16 cover-stego
pairs). The maximum numbers of iterations was set to 50 $\times$
10$^4$ (about 571 epochs) on BOSSBase + BOWS2 dataset, and 80 $\times$
10$^4$ (about 228 epochs) on ALASKA v2 dataset. The final searched
STD-NET models adopted the same maximum number of iterations as the
corresponding initial architecture. The optimizer used for the initial
architectures and STD-NET was Adamax with initial learning rate
0.001. After 40 $\times$ 10$^4$ iterations the learning rate was
reduced to 0.0001.  In our experiments, the cylinder-shaped,
ladder-shaped STD-NET architectures searched on JPEG-domain QF75
BOSSBase+BOWS2 and cylinder-shaped STD-NET architecture searched on
JPEG-domain QF75 ALASKA v2 are abbreviated as STD-BB-Cylinder,
STD-BB-Ladder and STD-ALA-Cylinder, respectively.

The one with the best validation accuracy was evaluated on the
corresponding testing set. All of the experiments were conducted on a
GPU cluster with single NVIDIA Tesla P100 GPU card. Bounded by
computational resources, every experiment was repeated three times,
and the mean of the results on testing set were reported.

The source codes and auxiliary materials are available for download
from GitHub.~\footnote{\url{https://github.com/tansq/STD-NET}}

\subsection{Determination of the normalized distortion values}
\label{sec:ma_determine}

As described in Sect.~\ref{sec:algorithm-criterion}, we need to determine the appropriate normalized distortion threshold $\mathscr{T}_d^l$ for every involved convolutional layer in the corresponding SRNetC64 model. In our experiments, the pre-defined step of shrinking rate was set to 5\%, and the pre-defined lower bound of the shrinking rate $\tau$ was set to 20\%. 
We start searching from the shrinking rate 50\% for the  bottom module, and 30\% for the middle module and ``L12'' of SRNetC64.

In Fig.~\ref{fig:ma_determine}\subref{fig:ma_determine_bb}, we show
how the normalized distortion $\mathscr{T}_\gamma$ changes with
successive decreasing shrinking rates on JPEG-domain QF75
BOSSBase+BOWS2. Similarly, the change curves on JPEG-domain QF75
ALASKA v2 datasets are shown in
Fig.~\ref{fig:ma_determine}\subref{fig:ma_determine_ala}.  It can be
seen that in most cases, as the shrinking rate decreases, the growth
rate of $\mathscr{T}_\gamma$ increases. This trend is particularly
obvious in the top module, and it also indicates that there is more
redundancy in the top module.  From Fig.~\ref{fig:ma_determine}, we
can see that the finally determined distortion threshold
$\mathscr{T}_d^l$ of the upper layers are generally smaller than those
of the lower layers.

\subsection{Compactness of STD-NET}
\label{sec:compactness}

In this section, we analyze the compactness and effectiveness of our proposed STD-NET.
Based on the determined normalized distortion, we follow Algorithm.~\ref{alg:input_output_channel_num_determination} to search for the corresponding number of input and output channels for the corresponding core tensor $\boldsymbol{\mathcal{G}}^{l}$.
Tab.~\ref{tab:std_detailed_structure} shows the three obtained structures, including
STD-BB-Cylinder, STD-BB-Ladder and STD--ALA-Cylinder.

Tab.~\ref{tab:std_params_flops} shows the comparison of the parameters
and FLOPs of the three different STD-NET structures with
CALPA-SRNet. Compared with CALPA-SRNet, our proposed STD-NET can
compress SRNet more effectively.  Parameters and FLOPs of the STD-NETs
are further reduced by about 70\% and 60\%, respectively, except for
the parameters of STD-ALA-Cylinder.  Specifically, all the involved
convolutional layers in STD-BB-Cylinder and STD-BB-Ladder get better
compression ratios compared with CALPA-NET, except those in ``L5'' and
``L10''. As for STD-ALA-Cylinder, it has more parameters than other
comparison models, since the ALASKA v2 dataset is more complex than
BOSSBase$+$BOWS2 and requires more model parameters to maintain
detection performance.

It is highlighted that compared with the original SRNet, parameters
and FLOPs of STD-BB-Cylinder, STD-BB-Ladder, and STD-ALA-Cylinder are
further reduced by 1.04\% and 20.16\%, 1.00\% and 19.18\%, and 1.76\%
and 22.38\%, respectively.



\begin{table*}[!h]
  \centering
  \caption[]{Comparison of detection performance of our proposed
    STD-NET under three different configurations and scenarios, and
    the corresponding CALPA-SRNet and the original SRNet. Datas with
    \pinkbox, \greenbox and \bluebox backgrounds are from a
    cylinder-shaped~(on BOSSBase+BOWS2), a ladder-shaped~(on
    BOSSBase+BOWS2) and a cylinder-shaped configuration~(on
    ALASKAv2/QF75/256$\times$256/gray-scale dataset), respectively.
    Datas with blue frames are from the corresponding STD-NET model
    following curriculum learning method.  }
  \adjustbox{center}{
    \resizebox{\textwidth}{!}{%
      {\renewcommand{\arraystretch}{1.3}
        \begin{tabular}{cc|cc|m{2.5em}<{\centering}m{7.5em}<{\centering}|m{3.5em}<{\centering}m{7.5em}<{\centering}|cc}
          \Xhline{2\arrayrulewidth}
          & & \multicolumn{2}{ c|}{} & \multicolumn{2}{ c| }{\textbf{STD-NETs for SRNet}} & \multicolumn{2}{ c| }{\textbf{CALPA-SRNet}} & \multicolumn{2}{ c }{\textbf{SRNet}} \\
          \hline
           & {\textbf{Scenarios}} & \multicolumn{2}{ c|}{\textbf{Targets}} & $P_{MD}(5\%)$ & $wAUC$ & $P_{MD}(5\%)$ & $wAUC$ & $P_{MD}(5\%)$  & $wAUC$ \\
          \hline
          \multirow{16}{*}{\rotatebox[origin=c]{90}{\textbf{BOSSBase+BOWS2}}} & \multirow{8}{*}{\makecell{\textbf{JPEG}\\\small(QF75)}} & \multirow{4}{*}{\textbf{J-UNIWARD}} & 0.4 bpnzAC &
            \noindent{\begin{tabular}{c}
                        \cellcolor{LightPink1}{\,10.9\%}\\
                        \cellcolor{DarkSeaGreen1}{\,12.4\%}\\
                      \end{tabular}} 
          &
            \noindent{\begin{tabular}{c}
                        \cellcolor{LightPink1}{\,98.3\%~\textbf{(\textdownarrow 0.4\%)}\,}\\
                        \cellcolor{DarkSeaGreen1}{\,98.4\%~\textbf{(\textdownarrow 0.3\%)}\,}\\
                      \end{tabular}} 
          & 10.3\% & 98.5\%~\textbf{(\textdownarrow 0.14\%)} & 9.80\% & 98.7\% \\
          & & & 0.2 bpnzAC &
                             \noindent{\begin{tabular}{c}
                                         \cellcolor{LightPink1}{\,47.8\%}\\
                                         \cellcolor{DarkSeaGreen1}{\,47.2\%}\\
                                       \end{tabular}} 
          &
            \noindent{\begin{tabular}{c}
                        \cellcolor{LightPink1}{\,91.1\%~\textbf{(\textdownarrow 1.2\%)}\,}\\
                        \cellcolor{DarkSeaGreen1}{\,90.7\%~\textbf{(\textdownarrow 1.6\%)}\,}\\
                      \end{tabular}} 
          & 43.9\% & 92.3\%~\textbf{(\textuparrow 0.04\%)} & 41.2\% & 92.3\% \\
          & & \multirow{4}{*}{\textbf{UERD}} & 0.4 bpnzAC &
                        \noindent{\begin{tabular}{c}
                                    \cellcolor{LightPink1}{\ \,1.6\%\,}\\
                                    \cellcolor{DarkSeaGreen1}{\ \,2.0\%\,}\\
                                  \end{tabular}} 
          &
            \noindent{\begin{tabular}{c}
                        \cellcolor{LightPink1}{\,99.7\%~\textbf{(\textdownarrow 0.0\%)}\,}\\
                        \cellcolor{DarkSeaGreen1}{\,99.7\%~\textbf{(\textdownarrow 0.0\%)}\,}\\
                      \end{tabular}} 
          & 1.9\% & 99.62\%~\textbf{(\textdownarrow 0.08\%)} & 1.7\% & 99.7\% \\
          & & & 0.2 bpnzAC &
                  \noindent{\begin{tabular}{c}
                              \cellcolor{LightPink1}{\,16.0\%}\\
                              \cellcolor{DarkSeaGreen1}{\,19.8\%}\\
                            \end{tabular}} 
          &
            \noindent{\begin{tabular}{c}
                        \cellcolor{LightPink1}{\,97.9\%~\textbf{(\textdownarrow 0.0\%)}\,}\\
                        \cellcolor{DarkSeaGreen1}{\,97.3\%~\textbf{(\textdownarrow 0.6\%)}\,}\\
                      \end{tabular}} 
          & 17.0\% & 97.7\%~\textbf{(\textdownarrow 0.2\%)} & 15.4\% & 97.9\% \\
          \cline{2-10}
          & \multirow{8}{*}{\textbf{Spatial}} & \multirow{4}{*}{\textbf{MiPOD}} & 0.4 bpp &
                              \noindent{\begin{tabular}{c}
                                          \cellcolor{LightPink1}{\,34.8\%}\\
                                          \cellcolor{DarkSeaGreen1}{\,33.0\%}\\
                                        \end{tabular}} 
          &
            \noindent{\begin{tabular}{c}
                        \cellcolor{LightPink1}{\,94.8\%~\textbf{(\textdownarrow 1.0\%)}\,}\\
                        \cellcolor{DarkSeaGreen1}{\,95.1\%~\textbf{(\textdownarrow 0.7\%)}\,}\\
                      \end{tabular}} 
          &  50.0\% &  94.4\%~\textbf{(\textdownarrow 1.4\%)} & 27.9\% & 95.8\% \\
          & & & 0.2 bpp &
            \noindent{\setlength\arrayrulewidth{2pt}\begin{tabular}{c}
                                                      \arrayrulecolor{blue}\hline
                                                      \multicolumn{1}{|c}{\cellcolor{LightPink1}{57.8\%}}\\
                                                      \multicolumn{1}{|c}{\cellcolor{DarkSeaGreen1}{57.4\%}}\\
                                                      \hline
                                                    \end{tabular}}
          &
            \noindent{\setlength\arrayrulewidth{2pt}\begin{tabular}{c}
                                                      \arrayrulecolor{blue}\hline
                                                      \multicolumn{1}{c|}{\cellcolor{LightPink1}{\,87.1\%~\textbf{(\textuparrow 1\%)}}\,}\\
                                                      \multicolumn{1}{c|}{\cellcolor{DarkSeaGreen1}{\,87.0\%~\textbf{(\textuparrow 0.9\%)}}\,}\\
                                                      \hline
                                                    \end{tabular}} 
          & 57.0\% & 86.0\%~\textbf{(\textdownarrow 0.1\%)} & 56.9\% & 86.1\% \\
          & & \multirow{4}{*}{\textbf{HILL}} & 0.4 bpp &
                      \noindent{\begin{tabular}{c}
                                  \cellcolor{LightPink1}{\,31.4\%}\\
                                  \cellcolor{DarkSeaGreen1}{\,30.6\%}\\
                                \end{tabular}} 
          &
            \noindent{\begin{tabular}{c}
                        \cellcolor{LightPink1}{\,95.3\%~\textbf{(\textdownarrow 0.7\%)}\,}\\
                        \cellcolor{DarkSeaGreen1}{\,95.8\%~\textbf{(\textdownarrow 0.2\%)}\,}\\
                      \end{tabular}} 
          & 28.0\% & 96.1\%~\textbf{(\textuparrow 0.13\%)} & 27.7\% & 96.0\% \\
          & & & 0.2 bpp &
                          \noindent{\begin{tabular}{c}
                                      \cellcolor{LightPink1}{\,53.8\%}\\
                                      \cellcolor{DarkSeaGreen1}{\,53.3\%}\\
                                    \end{tabular}} 
          &
            \noindent{\begin{tabular}{c}
                        \cellcolor{LightPink1}{\,88.6\%~\textbf{(\textdownarrow 1.1\%)}\,}\\
                        \cellcolor{DarkSeaGreen1}{\,89.3\%~\textbf{(\textdownarrow 0.4\%)}\,}\\
                      \end{tabular}} 
          & 49.4\% & 89.5\%~\textbf{(\textdownarrow 0.15\%)} & 50.6\% & 89.7\% \\
          \hline
          \multirow{16}{*}{\rotatebox[origin=c]{90}{\textbf{ALASKA v2}}} & \multirow{8}{*}{\makecell{\textbf{JPEG}\\\small(QF75)}} & \multirow{4}{*}{\textbf{J-UNIWARD}} & 0.4 bpnzAC &
                                                  \noindent{\begin{tabular}{c}
                                                              \cellcolor{LightPink1}{\,33.9\%}\\
                                                              \cellcolor{LightCyan2}{\,33.2\%}\\
                                                            \end{tabular}} 
          &
            \noindent{\begin{tabular}{c}
                        \cellcolor{LightPink1}{\,94.9\%~\textbf{(\textdownarrow 1.2\%)}\,}\\
                        \cellcolor{LightCyan2}{\,94.9\%~\textbf{(\textdownarrow 1.2\%)}\,}\\
                      \end{tabular}} 
          &  37.8\% &  93.1\%~\textbf{(\textdownarrow 3.0\%)} & 26.6\% & 96.1\% \\
          & & & 0.2 bpnzAC &
                             \noindent{\begin{tabular}{c}
                                         \cellcolor{LightPink1}{\,67.1\%}\\
                                         \cellcolor{LightCyan2}{\,68.8\%}\\
                                       \end{tabular}} 
          &
            \noindent{\begin{tabular}{c}
                        \cellcolor{LightPink1}{\,84.0\%~\textbf{(\textdownarrow 4.0\%)}\,}\\
                        \cellcolor{LightCyan2}{\,83.1\%~\textbf{(\textdownarrow 4.9\%)}\,}\\
                      \end{tabular}}
          &  76.5\% &  80.5\%~\textbf{(\textdownarrow 7.5\%)} & 84.1\% & 88.0\% \\
          & & \multirow{4}{*}{\textbf{UERD}} & 0.4 bpnzAC &
                        \noindent{\begin{tabular}{c}
                                    \cellcolor{LightPink1}{\,16.4\%}\\
                                    \cellcolor{LightCyan2}{\,20.5\%}\\
                                  \end{tabular}} 
          &
            \noindent{\begin{tabular}{c}
                        \cellcolor{LightPink1}{\,96.4\%~\textbf{(\textdownarrow 0.3\%)}\,}\\
                        \cellcolor{LightCyan2}{\,96.2\%~\textbf{(\textdownarrow 0.5\%)}\,}\\
                      \end{tabular}} 
          & 50.8\% & 92.8\%~\textbf{(\textdownarrow 3.9\%)} & 16.0\% & 96.7\% \\
          & & & 0.2 bpnzAC &
                             \noindent{\begin{tabular}{c}
                                         \cellcolor{LightPink1}{\,43.7\%}\\
                                         \cellcolor{LightCyan2}{\,47.9\%}\\
                                       \end{tabular}} 
          &
            \noindent{\begin{tabular}{c}
                        \cellcolor{LightPink1}{\,92.5\%~\textbf{(\textuparrow 1.5\%)}\,}\\
                        \cellcolor{LightCyan2}{\,90.5\%~\textbf{(\textdownarrow 0.5\%)}\,}\\
                      \end{tabular}} 
          & 65.3\% & 72.1\%~\textbf{(\textdownarrow 18.9\%)} & 45.0\% & 91.0\% \\
          \cline{2-10}
          & \multirow{4}{*}{\makecell{\textbf{JPEG}\\\small(various)}} & \multirow{4}{*}{\textbf{J-UNIWARD}} & 0.4 bpnzAC &
                                    \noindent{\begin{tabular}{c}
                                                \cellcolor{LightPink1}{\,64.5\%}\\
                                                \cellcolor{LightCyan2}{\,65.3\%}\\
                                              \end{tabular}} 
          &
            \noindent{\begin{tabular}{c}
                        \cellcolor{LightPink1}{\,84.8\%~\textbf{(\textuparrow 0.3\%)}\,}\\
                        \cellcolor{LightCyan2}{\,84.4\%~\textbf{(\textdownarrow 0.1\%)}\,}\\
                      \end{tabular}} 
          & 73.5\% & 83.2\%~\textbf{(\textdownarrow 1.3\%)} & 65.0\% & 84.5\% \\
          & & & 0.2 bpnzAC &
                             \noindent{\setlength\arrayrulewidth{2pt}\begin{tabular}{c}
                                                  \arrayrulecolor{blue}\hline
                                                  \multicolumn{1}{|c}{\cellcolor{LightPink1}{86.1\%}}\\
                                                  \hline
                                                  \cellcolor{LightCyan2}{93.6\%}\\
                                                \end{tabular}} 
          &
            \noindent{\setlength\arrayrulewidth{2pt}\begin{tabular}{c}
                                                      \arrayrulecolor{blue}\hline
                                                      \multicolumn{1}{c|}{\cellcolor{LightPink1}{\,71.7\%~\textbf{(\textuparrow 1.7\%)}}\,}\\
                                                      \hline
                                                      \cellcolor{LightCyan2}{61.6\%~\textbf{(\textdownarrow 8.4\%)}}\\
                                                    \end{tabular}} 
          &  93.1\% &  65.2\%~\textbf{(\textdownarrow 4.8\%)} & 89.6\% & 70.0\% \\
          & & \multirow{2}{*}{\textbf{UERD}} & 0.4 bpnzAC &
                                                            \noindent{\begin{tabular}{c}
                                                                        \cellcolor{LightPink1}{\,50.1\%}\\
                                                                        \cellcolor{LightCyan2}{\,49.1\%}\\
                                                                      \end{tabular}} 
          &
            \noindent{\begin{tabular}{c}
                        \cellcolor{LightPink1}{\,90.5\%~\textbf{(\textdownarrow 0.5\%)}}\\
                        \cellcolor{LightCyan2}{\,89.6\%~\textbf{(\textdownarrow 1.4\%)}\,}\\
                      \end{tabular}} 
          &  57.7\% &  87.0\%~\textbf{(\textdownarrow 4.0\%)} & 49.0\% & 91.0\% \\
          & & & 0.2 bpnzAC &
            \noindent{\setlength\arrayrulewidth{2pt}\begin{tabular}{c}
                                                      \arrayrulecolor{blue}
                                                      \hline
                                                      \multicolumn{1}{|c}{\cellcolor{LightPink1}{71.5\%}}\\
                                                      \hline
                                                      \cellcolor{LightCyan2}{93.3\%}\\
                                                    \end{tabular}} 
          &
            \noindent{\setlength\arrayrulewidth{2pt}\begin{tabular}{c}
                                                      \arrayrulecolor{blue}
                                                      \hline
                                                      \multicolumn{1}{c|}{\cellcolor{LightPink1}{80.9\%~\textbf{(\textuparrow 0.9\%)}}}\\
                                                      \hline
                                                      \cellcolor{LightCyan2}{61.4\%~\textbf{(\textdownarrow 18.6\%)}}\\
                                                    \end{tabular}} 
          &  87.6\% & 75.0\%~\textbf{(\textdownarrow 5.0\%)} & 80.1\% & 80.0\% \\
          \Xhline{2\arrayrulewidth}
        \end{tabular} 
      }
    }
  }
  \label{tab:std_comp_state_of_the_art}
\end{table*}

\subsection{Detection performance of STD-NETs}
\label{sec:exp_detection}


In Tab.~\ref{tab:std_comp_state_of_the_art}, we compare the detection performance of the three STD-NETs, the corresponding CALPA-SRNet and the original SRNet on eighteen different scenarios.

In the JPEG domain, the detection performance of the STD-BB-Cylinder
and STD-BB-Ladder on almost all involved scenarios outperforms
CALPA-SRNet.  In particular, the cylinder-shaped STD-NETs even exceed the original SRNet on
JPEG-domain  ALASKA v2 dataset with various quality factors.

In the spatial domain, the two variants of STD-NETs (STD-BB-Cylinder
and STD-BB-Ladder), originally searched on JPEG domain dataset (QF75
BOSSBase+BOWS2 aiming at J-UNIWARD 0.4 bpnzAC), still achieve similar
or even better detection performance compared with CALPA-SRNet, as
well as original SRNet.

In the following experiments, unless otherwise specified, we only
report the results for those STD-NET models with the structures
obtained on the JPEG-domain QF75 BOSSBase+BOWS2 dataset.

\begin{figure*}[!t]
  \centering
  \subfloat[]{
    \label{fig:acc_vs_iter_srnet_bbqf75ju04}  
    \includegraphics[width=0.32\textwidth,keepaspectratio]{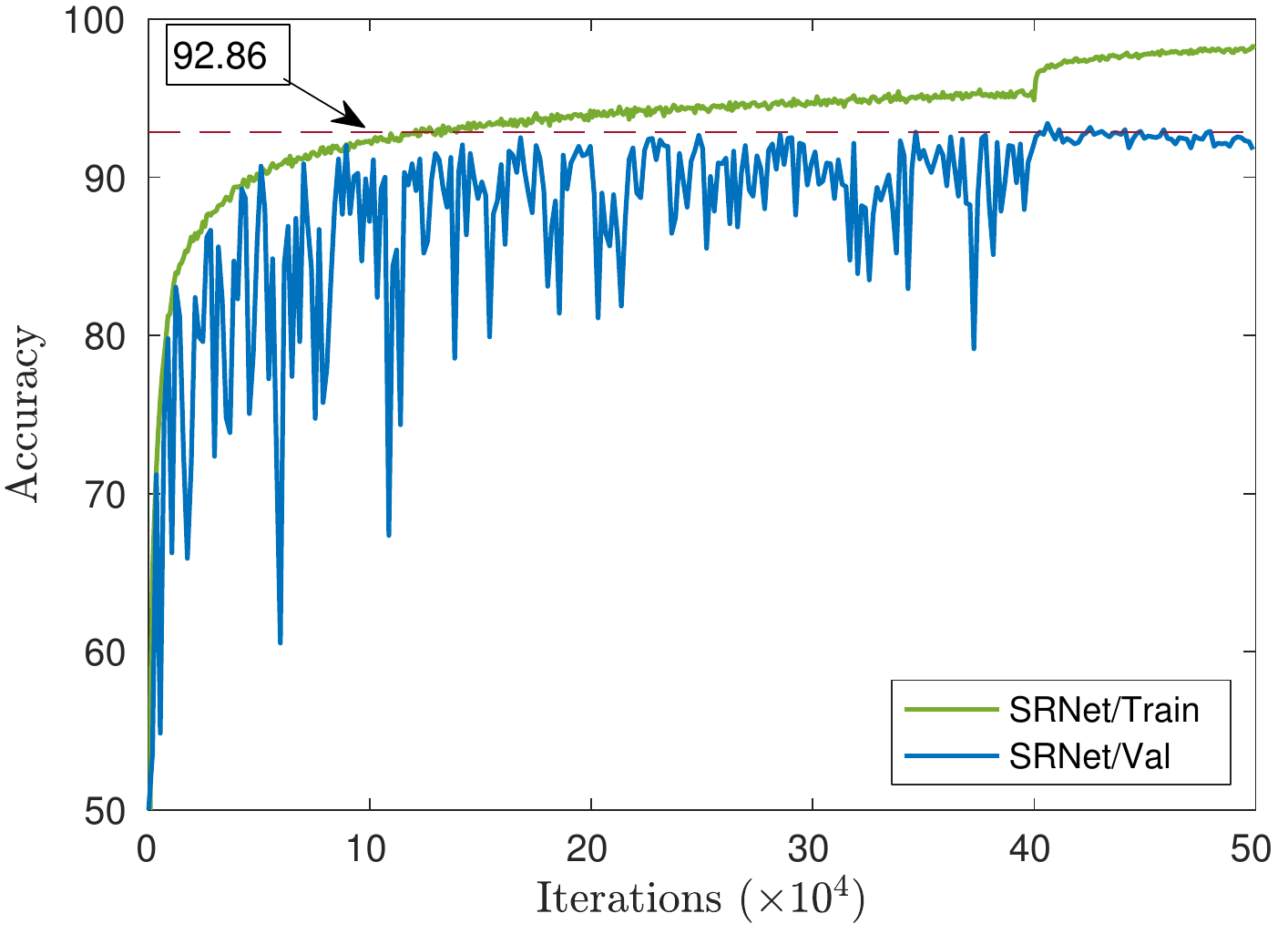}
  }
  \subfloat[]{
    \label{fig:acc_vs_iter_bbv1_bbqf75ju04}  
    \includegraphics[width=0.32\textwidth,keepaspectratio]{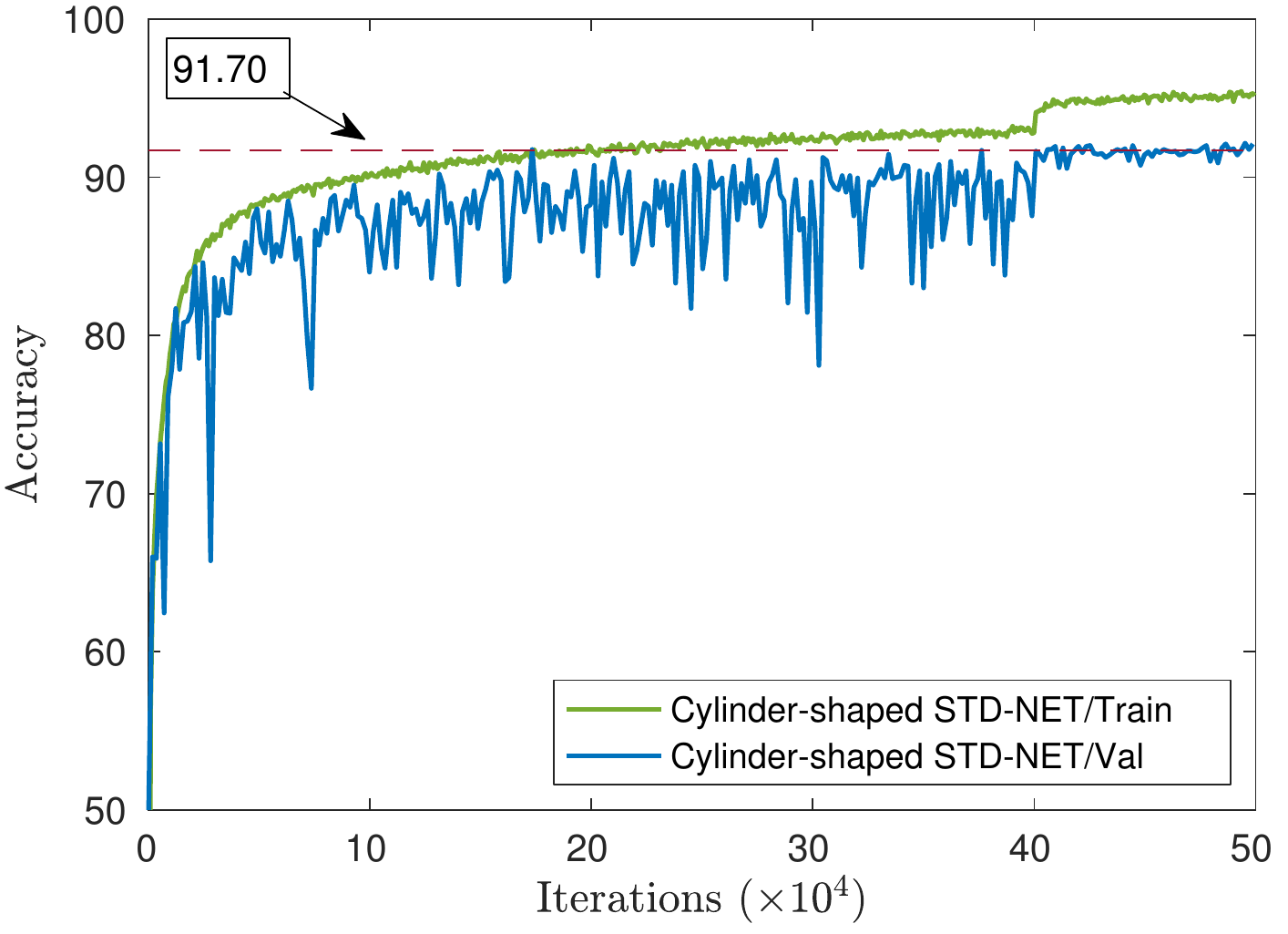}
  }
  \subfloat[]{
    \label{fig:acc_vs_iter_bbv3_bbqf75ju04}  
    \includegraphics[width=0.32\textwidth,keepaspectratio]{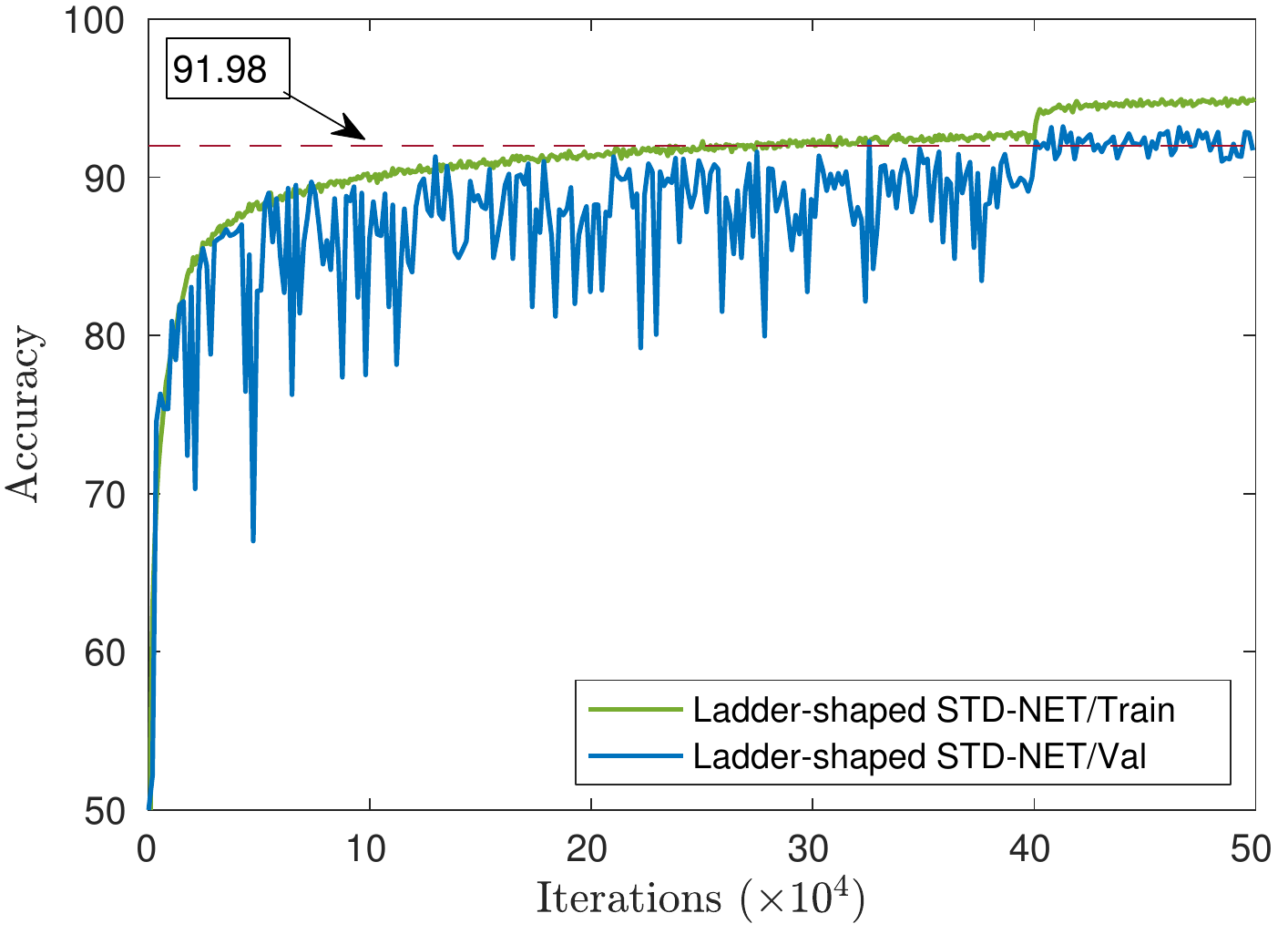}
  }\\
  \caption[]{ Comparison of the training accuracies and validation
    accuracies vs. training iterations for SRNet and the corresponding
    STD-NETs~(STD-BB-Cylinder and STD-BB-Ladder). The dashed line in
    every sub-figure marks the testing accuracy achieved by the
    trained model with highest validation
    accuracy. \subref{fig:acc_vs_iter_srnet_bbqf75ju04},
    \subref{fig:acc_vs_iter_bbv1_bbqf75ju04}, and 
    \subref{fig:acc_vs_iter_bbv3_bbqf75ju04} is the result for
    original SRNet, cylinder-shaped STD-NET, and ladder-shaped
    STD-NET, respectively. The experiments were conducted on
    JPEG-domain QF75 BOSSBase+BOWS2 dataset, aiming at J-UNIWARD/0.4
    bpnzAC.}
  \label{fig:acc_vs_iter}  
\end{figure*}

Fig.~\ref{fig:acc_vs_iter} shows the performance comparison of the
STD-BB-Cylinder, STD-BB-Ladder STD-NETs and original SRNet, on
JPEG-domain QF75 BOSSBase+BOWS2 dataset aiming at J-UNIWARD/0.4
bpnzAC.  We can see that the cylinder and ladder-shaped STD-NETs show
similar validation and testing performance compared with original
SRNet.  It is worth noting that the larger the model is, the bigger
the gap between training accuracies and validation accuracies for the
corresponding model is.  The gaps of SRNet, STD-BB-Cylinder, and
STD-BB-Ladder are clearly reduced in accordance with their model
scale, which has been sorted from the largest to the smallest.


From Tab.~\ref{tab:std_comp_state_of_the_art} and
Fig.~\ref{fig:acc_vs_iter}, we can see our STD-NET architectures,
especially STD-BB-Cylinder and STD-BB-Ladder, can obtain competitive
detection performance in most scenarios. For instance, compared with
original SRNet, the two STD-NET architectures can achieve competitive
accuracy even if it is trained from scratch (see
Fig.~\ref{fig:acc_vs_iter}). Therefore, we can make a conclusion that
our STD-NET has a great adaptivity.

\begin{figure*}[!t]
	\centering
        \includegraphics[width=0.8\textwidth,keepaspectratio]{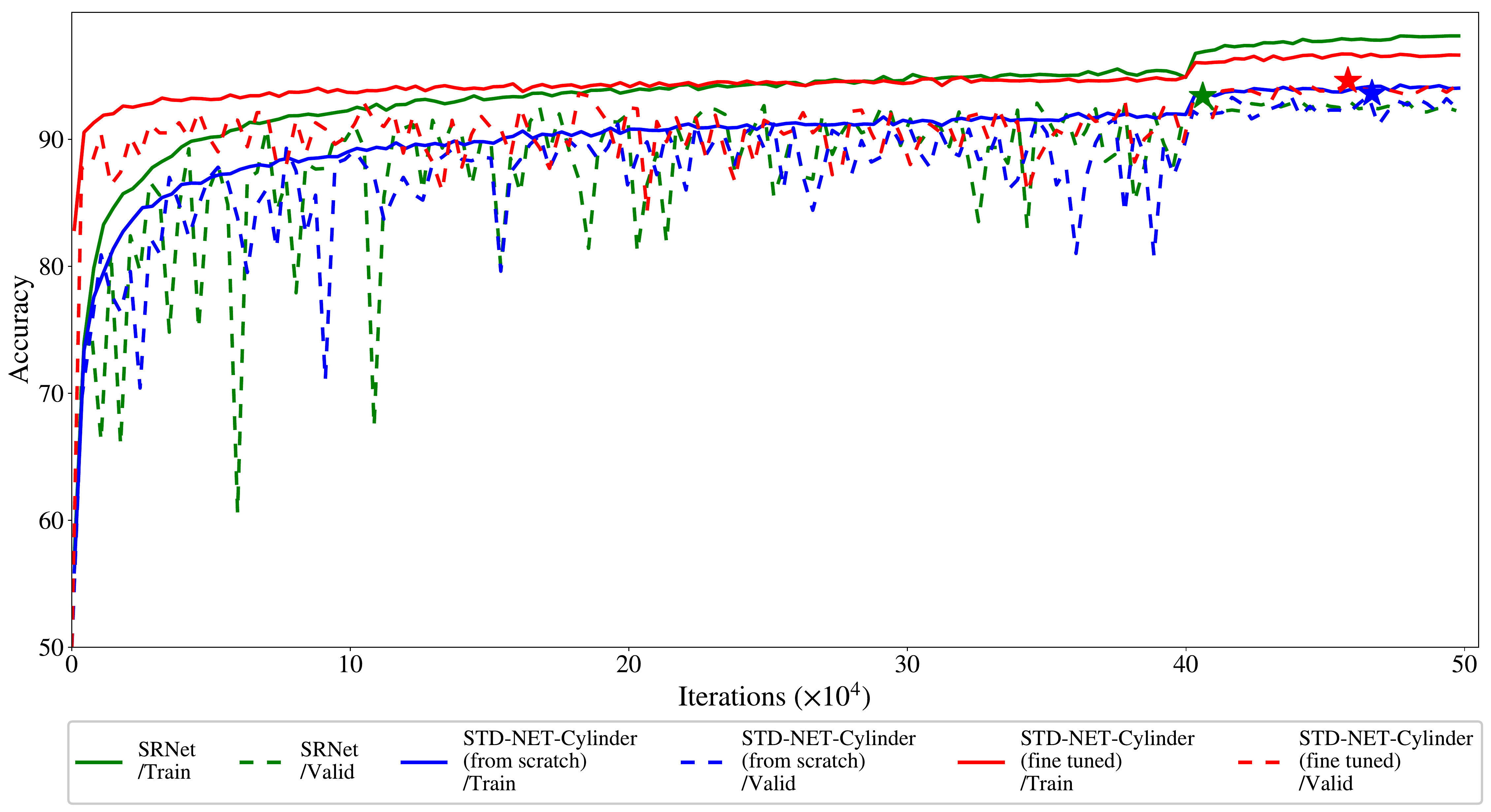}

        \caption[]{ Comparison of the training accuracies and
          validation accuracies vs. training iterations for the
          original SRNet, corresponding STD-NET trained from scratch,
          and STD-NET with ``training-decomposing-finetuning''
          pipeline. The experiment was conducted on BOSSBase+BOWS2
          dataset aiming at J-UNIWARD/0.4 bpnzAC. ``\FiveStar'' denotes
          the point with highest value at the corresponding plot.}
        \label{fig:comp_fine_tuned_from_scratch_bbqf75ju04}
\end{figure*}

Fig.~\ref{fig:comp_fine_tuned_from_scratch_bbqf75ju04} shows
comparison of training accuracies and validation accuracies
vs. training iterations for the original SRNet and two STD-NET models
with two different training strategy. From
Fig.~\ref{fig:comp_fine_tuned_from_scratch_bbqf75ju04} we can see that
both training strategy, namely ``training-decomposing-finetuning'' and
``trained-from-scratch'', guarantee the STD-NET models achieve similar
performance when compared with SRNet on validation set.  The STD-NET
model with ``training-decomposing-finetuning'' pipeline converges
faster than other models on the training set.

All in all, on one hand, because our STD-NET is an adaptive
architecture, we can directly train from scratch without using an
existing model. On the other hand, the STD-NET model that preserves
the model parameters after Tucker decomposition has a certain
expression and discriminative ability, which helps to converge faster
on varying steganalysis tasks.

\subsection{Detection performance on non-additive steganography}
\label{sec:non_additive}

To further investigate our STD-NET models' detection ability,  we evaluate its detection performance on BOSSBase+BOWS2, targeting at non-additive steganography. In spatial domain, we adopt HILL as basic additive cost function with the state-of-the-art non-additive steganographic framework CMD~\cite{li_tifs_2015} at payload of 0.4 bpp. In JPEG domain, J-UNIWARD is chosen as basic additive steganography with non-additive steganographic frameworks BBM~\cite{wang_tifs_2021_bbm} and BBC-BBM (the combination of BBC~\cite{li_ihmmsec_2018} and BBM).

In Tab.~\ref{tab:non_additive}, we compare the detection performance of the STD-NET models STD-BB-Cylinder and STD-BB-Ladder with the original SRNet.
Please note the two STD-NET models' parameters are finetuned after Tucker decomposition. 
From Tab.~\ref{tab:non_additive} we can see the fine-tuned STD-NET model gets similar or even better performance compared with SRNet.
The performance of the two models, STD-BB-Cylinder and STD-BB-Ladder, is very close. Compared with SRNet, their accuracy losses do not exceed 0.2\%. Further on, targeting at BBM, the performance of our two STD-NET models is even 0.11\% and 0.05\% higher than that of SRNet model, respectively.

\begin{table*}[!t]
  \centering
  \caption[]{Comparison of detection accuracy for SRNet and the
    corresponding STD-NET models (with the
    ``training-decomposing-finetuning'' pipeline) searched on
    JPEG-domain QF75 BOSSBase+BOWS2 at non-additive steganography.}
  \adjustbox{center}{
    \setlength{\tabcolsep}{12pt}.	
    \renewcommand{\arraystretch}{1.3}	
    \begin{tabular}{$c^c^c^c}
      \Xhline{3\arrayrulewidth}
      \multirow{2}{*}{Model} & HILL 0.4 bpp & \multicolumn{2}{c}{J-UNIWARD 0.4 bpnzAC} \\
      \cmidrule(r){2-2}  \cmidrule(r){3-4}
                             & CMD & BBM & BBC-BBM \\
      \Xhline{2\arrayrulewidth}
      SRNet & 81.99 & 91.63 & 91.40 \\
      \hline
      STD-BB-Cylinder (fine-tuned) & 81.92 ($\downarrow$0.07) & 91.74 ($\uparrow$0.11) & 91.22 ($\downarrow$0.18) \\
      STD-BB-Ladder (fine-tuned) & 81.94 ($\downarrow$0.05) & 91.68 ($\uparrow$0.05) & 91.20 ($\downarrow$0.20) \\		
      \Xhline{3\arrayrulewidth}
    \end{tabular}
  }
  \label{tab:non_additive}
\end{table*}

\begin{table*}[!t]
  \centering
  \caption[]{The comparison of model size and performance before and after decomposition for SiaStegNet through our proposed STD-NET .}
  \adjustbox{center}{
    \setlength{\tabcolsep}{12pt}.	
    \renewcommand{\arraystretch}{1.3}	
    \begin{tabular}{$c^c^c^c}
      \Xhline{3\arrayrulewidth}
      Model & Params (M) & FLOPs (M) & ACC (\%) \\
      \Xhline{2\arrayrulewidth}
      SiaStegNet & 1.41 & 7159.9 & 88.83 \\
      \hline
      STD-Sia-Cylinder (fine-tuned) & 0.25 (17.86\%) & 1874.46 (26.17\%) & 89.42 ($\uparrow$0.59) \\
      STD-Sia-Cylinder (from scratch) & 0.25 (17.86\%) & 1874.46 (26.17\%) & 88.58 ($\downarrow$0.25) \\
      STD-Sia-Ladder (fine-tuned) & 0.22 (16.01\%) & 1739.79 (24.29\%) & 89.32 ($\uparrow$0.49) \\		
      STD-Sia-Ladder (from scratch) & 0.22 (16.01\%) & 1739.79 (24.29\%) & 86.84 ($\downarrow$1.99) \\		
      \Xhline{3\arrayrulewidth}
    \end{tabular}
  }
  \label{tab:decom_othernet}
\end{table*}

\begin{table*}[!h]
  \centering
  \caption[]{Comparison of detection performance for SRNet and the
    corresponding STD-NET models (with the ``train-from-scratch''
    pipeline) aiming at J-UNIWARD 0.4 bpnzAC. The dataset is
    BOSSBase+BOWS2. The cylinder-shaped
    as well as the ladder-shaped STD-NETs are searched on the
    corresponding scenarios.}
  \adjustbox{center}{
    \resizebox{0.6\textwidth}{!}{%
    \renewcommand{\arraystretch}{1.3}	
    \begin{tabular}{$c^c^c^c^c}
      \Xhline{3\arrayrulewidth}
      \multirow{2}{*}{Model} & \multicolumn{2}{c}{QF90/$256 \times
                               256$} & \multicolumn{2}{c}{QF75/$512
                                       \times 512$}  \\
      \cmidrule(r){2-3}  \cmidrule(r){4-5}
                             &  $P_{MD}(5\%)$ & $wAUC$ & $P_{MD}(5\%)$ & $wAUC$  \\
      \Xhline{2\arrayrulewidth}
      SRNet & 20.0 & 97.4 & 9.8  & 98.1   \\
      \hline
      CALPA-NET & 24.60 & 96.73~\textbf{(\textdownarrow 0.67)}  & 9.36 & 97.69~\textbf{(\textdownarrow 0.41)}   \\
      \hline
      \makecell{STD-NET\\cylinder-shaped}  & 21.91& 96.94~\textbf{(\textdownarrow 0.46)} & 5.4 & 98.5~\textbf{(\textuparrow 0.4)}   \\
      \makecell{STD-NET\\ladder-shaped} & 21.24 & 96.97~\textbf{(\textdownarrow 0.43)} & 13.7  & 96.6~\textbf{(\textdownarrow 1.5)} \\
      STD-BB-FIXED & 30.2 & 95.4~\textbf{(\textdownarrow 2)} & 13.6 & 97.2~\textbf{(\textdownarrow 0.9)}        \\
      \Xhline{3\arrayrulewidth}
    \end{tabular}
  }
  }
\label{tab:std_comp_90_512}
\end{table*}
  
\subsection{Generalization of STD-NET}
\label{sec:non_additive}

To evaluate the generalization of STD-NET, we further implement our
STD-NET strategy for SiaStegNet~\cite{you_tifs_2021}.  We trained the
original network and the searched versions on BOSSBase + BOWS2 dataset
aiming at S-UNIWARD 0.4 bpp for 500 epochs with batch size 32 (namely
16 cover-stego pairs). The optimizer used for the original
architecture and its corresponding STD-NET versions was Adamax with
initial learning rate 0.001 and weight decay 0.0001. All other
hyperparameters and the training schedule were same as described in
\cite{you_tifs_2021}. The two final obtained architectures are denoted
as STD-Sia-Cylinder and STD-Sia-Ladder, standing for the
cylinder-shaped and ladder-shaped STD-NET version of SiaStegNet,
respectively.

The comparison results of model size and performance before and after
decomposition are shown in Tab.~\ref{tab:decom_othernet}.  It can be
seen that the number of parameters and FLOPs of STD-Sia-Cylinder and
STD-Sia-Ladder are around 17\% and 25\% of the original respectively.
Those fine-tuned STD-NET models with decomposed parameters retained
can even gain better performance compared with the original SiaStegNet
model. Besides, though those STD-NET models trained from scratch
suffer losses of detection accuracy, the losses are still within a
tolerable range.

Generally speaking, our proposed STD-NET approach is available for all
the deep-learning steganalyzers equipped with convolutional layers,
and its applicative scenarios are rich. Besides the results dealing
with QF75 JPEG images, we have demonstrated its effectiveness in
following scenarios:
\begin{itemize}
\item JPEG-domain QF90 $256 \times 256$ scenario;
\item JPEG-domain QF75 $512 \times 512$ scenario.
\end{itemize}
As shown in Tab.~\ref{tab:std_comp_90_512}, even in the above two
harder scenarios, the detection performance of the cylinder-shaped as
well as the ladder-shaped STD-NETs searched on those scenarios can be
close to the original SRNet, and outperforms CALPA-SRNet.

We have compared our proposed unsupervised data-driven criterion with
manually selected core-tensor ranks used in vanilla Tucker
decomposition. We name the obtained cylinder-shaped STD-NET
architecture for JPEG-domain QF75 BOSSBase +BOWS2 aiming at
J-UNIWARD/0.4 bpnzAC, with vanilla Tucker decomposition as
STD-BB-FIXED, in which the shrinking rate for the bottom module is
fixed to 50\%, and the shrinking rate for the middle module and
``L12'' is fixed to 30\%.

As shown in Fig.~\ref{tab:std_comp_90_512}, It is no doubt that the
cylinder-shaped as well as the ladder-shaped STD-NETs searched on
those scenarios outperforms the corresponding STD-NETs with vanilla
Tucker decomposition.
  \begin{table*}[!t]
    \centering
    \caption[]{Comparison of detection performance for SRNet and the
      corresponding STD-NET models (with the ``train-from-scratch''
      pipeline searched on JPEG-domain QF75 BOSSBase+BOWS2) without
      pair constraint in the training procedure. The results for
      ImageNet pretrained EfficientNetB0 are reported as well. The
      underlined results are obtained with curriculum learning.}
    \adjustbox{center}{
      \resizebox{0.9\textwidth}{!}{%
      \renewcommand{\arraystretch}{1.3}	
      \begin{tabular}{$c^c^c^c^c^c^c^c^c}
        \Xhline{3\arrayrulewidth}
        \multirow{2}{*}{Model} & \multicolumn{2}{c}{BB-QF75-JU04} & \multicolumn{2}{c}{BB-QF75-UE04} 
        & \multicolumn{2}{c}{BB-QF90-JU04} & \multicolumn{2}{c}{ALA2-QF75-JU04} \\
        \cmidrule(r){2-3}  \cmidrule(r){4-5}  \cmidrule(r){6-7} \cmidrule(r){8-9}
                               & $P_{MD}(5\%)$ & $wAUC$  & $P_{MD}(5\%)$ & $wAUC$ & $P_{MD}(5\%)$ & $wAUC$ & $P_{MD}(5\%)$ & $wAUC$   \\
        \Xhline{2\arrayrulewidth}
        SRNet & 12.16 & 98.57 & 2.60   & 99.64 & 25.08 & 96.71 & 26.60 & 96.10 \\
        \hline
        EfficientNet-B0 & 66.6 & 86.63 & 58.20 & 92.10  & \underline{86.26} & \underline{75.46} & 29.01 & 95.12 \\
        \hline
        STD-BB-Cylinder & 10.84 & 98.53 & 2.11 & 99.66 & 29.20 & 96.01 & 33.91 & 94.91 \\
        STD-BB-Ladder  & 16.55 & 97.89 & 2.00 & 99.63 & 35.32 & 94.52 & 33.2 & 94.93 \\
        \Xhline{3\arrayrulewidth}
      \end{tabular}
    } 
    }
    \label{tab:comparison_nopc}
  \end{table*}
\subsection{Further discussions}
\label{sec:discussion}

\subsubsection{Impact of no pair constraint}
\label{sec:impact-no-pair}

Experimental results reported in
\cite{yousfi_wifs_2020,butora_ihmmsec_2021,yousfi_ih_2021} indicate
that training with pair constraint removed in the later stage may
improve the detection performance. An experiment has been conducted on
BOSSBase+BOWS2 dataset as well as ALASKA v2 dataset to verify the
impact of no pair constraint on the performance of our proposed
STD-NET approach. In the experiment all the involved models were
trained with 300 pair-constraint epochs and then 100
no-pair-constraint epochs. The results for a ImageNet pretrained
EfficientNet-B0 model are reported as well.

compared the results in Tab.~\ref{tab:comparison_nopc} with those in
Tab.~\ref{tab:std_comp_state_of_the_art}, we can see that on
BOSSBase+BOWS2 dataset training without pair constraint does actually
not introduce any relative advantage for the more complex SRNet
models. Compared with the results reported in Tab.III of the revision,
additional no-pair-constraint training epochs do mildly degrade the
detection performance of SRNet models, as well as our proposed STD-NET
models. Anyhow, on BOSSBase+BOWS2 dataset the extent of the impact of
no-pair-constraint training on SRNet and our proposed STD-NET is
similar. Since no obvious impact can be observed, we have sticked to
using the more straightforward pair-constraint training on
BOSSBase+BOWS2 dataset, , and reported obtained results in
Tab.~\ref{tab:std_comp_state_of_the_art}.

On the contrary, in the experiments we have observed that pair
constraint does introduce obvious detection performance improvements
for SRNet models on ALASKA v2 dataset. Now the huge gap between
training and validation for SRNet models trained with pair constraint
is markedly narrowed. Accordingly, we have removed pair constraint in
the later stage of the training procedure on ALASKA v2 dataset, and
report those  obtained results in Tab.~\ref{tab:std_comp_state_of_the_art}.

However, please note that as reported in
Tab.~\ref{tab:std_comp_state_of_the_art}, on ALASKA v2 dataset, even
with pair constraint removed in the later stage of the training
procedure, by and large the detection performances of our proposed
STD-NETs are still comparable to those of the original SRNets.
  
\subsubsection{STD-NET and EfficientNet family}
\label{sec:std-net-efficientnet}

Please note that though as reported in \cite{yousfi_ih_2021}, the
EfficientNet family~\cite{tan_icml_2019} pre-trained on ImageNet can
achieve state-of-the-art performance for JPEG steganalysis, STD-NET
cannot be directly applied to EfficientNet to effectively shrink
it. This is due to the fact that EfficientNets are mainly composed of
\textit{depthwise separable convolutional layers}, which cannot be
handled with existing tensor decomposition techniques. Depthwise
separable convolution is one of the most important components of EfficientNet
family. However, depthwise separable convolutions sharply reduce the
complexity margin which Tucker decomposition can cut off.

Given a block with depthwise separable convolution, it takes an input
tensor
$\boldsymbol{\mathcal{I}}^{l-1} \in \mathbb{R}^{J^{l-1} \times H^{l-1}
  \times W^{l-1}}$ and maps it into an output tensor
$\boldsymbol{\mathcal{O}}^{l} \in \mathbb{R}^{J^{l} \times H^{l}
  \times W^{l}}$ with a depthwise convolution followed by a pointwise
$1 \times 1$ convolution. Let
$\uunderset{d}{\boldsymbol{\mathcal{K}}}^{l} \in \mathbb{R}^{J^{l-1}
  \times D^{l} \times D^{l}}$ denotes the depthwise convolution kernel
tensor and
$\uunderset{p}{\boldsymbol{\mathcal{K}}}^{l} \in \mathbb{R}^{J^{l-1}
  \times J^{l}}$ denotes the pointwise convolution kernel
tensor. Elementwise~(definitions of $h_k$ and $w_s$ are the same as
those in Eq.~\eqref{eq:conv}, and are omitted here for clarity):
\begin{equation}
  \label{eq:ds_conv}
  \boldsymbol{\mathcal{O}}_{i,h,w}^{l}=\underbrace{\sum_{j=1}^{J^{l-1}}\uunderset{p}{\boldsymbol{\mathcal{K}}}_{\raisemath{1ex}{j,i}}^{l}\underbrace{\sum_{k=1}^{D^l}\sum_{s=1}^{D^l}\uunderset{d}{\boldsymbol{\mathcal{K}}}_{\raisemath{1ex}{j,k,s}}^{l} \cdot \boldsymbol{\mathcal{I}}_{j,h_k,w_s}^{l-1}}_{\textrm{depthwise conv}}}_{\textrm{pointwise conv}}\
\end{equation}
Measured with the number of parameters, the computational complexity of the
block turns to:
\begin{equation}
  \label{eq:ds_ll_flops}
  J^{l-1} \cdot D^{l} \cdot D^{l} + J^{l-1} \cdot J^{l}
\end{equation}
Assuming that we apply Tucker decomposition to
$\uunderset{d}{\boldsymbol{\mathcal{K}}}^{l}$, the resulting
complexity turns to:
\begin{equation}
  \label{eq:ds_ll_flops_after_tucker}
  J^{l-1}\cdot I^{l}+I^{l} \cdot O^{l} \cdot D^{l} \cdot D^{l}+O^{l} + J^{l-1} \cdot J^{l}
\end{equation}
Then subtract Eq.~\eqref{eq:ds_ll_flops_after_tucker} from
Eq.~\eqref{eq:ds_ll_flops}, we turn to solve the following inequality:
\begin{equation}
  \label{eq:ds_complexity_tucker_inequality}
  J^{l-1} \cdot D^{l} \cdot D^{l}-J^{l-1}\cdot I^{l}-I^{l} \cdot O^{l} \cdot D^{l} \cdot D^{l}-O^{l}\ge 0
\end{equation}
Again we set $D^{l} \cdot D^{l}=n$, $J^{l-1} = J^{l}=x$, and
$I^{l}=O^{l}=x\gamma,\ \gamma \in
[0,1]$. Eq.~\eqref{eq:ds_complexity_tucker_inequality} turns to:
\begin{equation}
  \label{eq:ds_complexity_tucker_inequality_deduction}
  nx-x^2\gamma  - n x^2\gamma^2-x \gamma \ge 0 \Rightarrow
  nx\gamma^2+(x+1)\gamma -n \le 0
\end{equation}
Regarding the left side of
Eq.~\eqref{eq:ds_complexity_tucker_inequality_deduction} as a
univariate quadratic function of $\gamma$, we can get that
Eq.~\eqref{eq:ds_complexity_tucker_inequality_deduction} holds in
$\gamma \in [0, \frac{-(x+1)+\sqrt{(x+1)^2+4n^2x}}{2nx}]$.  Since
$\sqrt{(x+1)^2+4n^2x} \le (x+1)+2n\sqrt{x}$, we can get
$\frac{-(x+1)+\sqrt{(x+1)^2+4n^2x}}{2nx} \le \frac{1}{\sqrt{x}}$. As a
result, $\gamma$ actually lies in a quite narrow range included in
$[0, \frac{1}{\sqrt{x}}]$. For instance, with $n=3 \times 3=9$ and
$x=64$, Tucker decomposition can only bring in computational
complexity reduction when the shrinking rate $\gamma < 8.1\%$.  Since
EfficientNet family has been well designed to be very compact given a
number of parameters, With such an aggressive shrinking rate, Tucker
decomposition will certainly severely degrade the performance of the
resulting model.

However, according to the experimental results reported in
\cite{yousfi_wifs_2020,butora_ihmmsec_2021,yousfi_ih_2021}, the
EfficientNet family is still far from the best framework for image
steganalysis.  Firstly, it even cannot converge without pre-trained
model weights obtained in other large-scale tasks. Secondly, even with
pre-trained model weights, vanilla EfficientNet is inferior to SRNet
in gray-scale image steganalysis as well as in aggressively
down-sampled images. Thirdly, though designed for compactness,
EfficientNet family is actually not superior in complexity even
compared with SRNet, not to mention our proposed STD-NETs.

The experimental results in Tab.~\ref{tab:comparison_nopc} provide
additional evidences. Tab.~\ref{tab:comparison_nopc} shows that the
ImageNet pretrained EfficientNet-B0 model is inferior even with
additional no-pair-constraint training epochs on BOSSBase+BOWS2
dataset. On ALASKA v2 dataset its performance cannot surpass that of
SRNet. Certainly, the reported results are obtained for vanilla
EfficientNet on gray-scale ALASKA v2 images. As reported in
\cite{yousfi_ih_2021}, specific renovations can improve the detection
performance of EfficientNet in image steganalysis. However, all of
those renovations increase its complexity by a wide margin.

To summarize, in a laboratory setting state-of-the-art deep-learning
steganalyzers can detect stego images with satisfactory accuracies in
quite a few scenarios. Therefore we are well past the stage that the
scholars were only chasing performance improvements. Besides detection
performance, computational complexity is another dimension to measure
the overall virtue of a deep-learning steganalyzer. Our proposed
STD-NET approach can achieve state-of-the-art performance on top of
various notable architectures with even an order of magnitude smaller
complexity. It can be highly complementary to the EfficientNet family
in the field of image steganalysis.

\section{Concluding remarks}
\label{sec:conclude}


In this paper, we propose STD-NET, which is aiming at searching an
efficient image steganalytic deep-learning architecture to save the
memory cost as well as the computational cost. The major contributions
of this work are as follows:
\begin{itemize}
\item We have proposed a hierarchical tensor decomposition strategy, which can greatly reduce model parameters and FLOPs through Tucker decomposition.
  Unlike CALPA-NET, the STD-NET will not be restricted by various residual shortcut connections.

\item We have proposed a normalized distortion threshold, which 
  guide us to decompose involved convolutional layers on the basis of the original SRNet model in an unsupervised way, so as to search for an efficient and adaptive deep-learning image steganalysis architecture.

\item The extensive experiments conducted on de-facto benchmarking image datasets show that 
  our STD-NET models achieve comparative detection performance whether it is finetuned with the decomposed parameters or trained from scratch.
\end{itemize}

In the future, we will mainly focus on two aspects: 1) explore fully
automatic deep steganalytic neural architecture search strategies; 2)
explore broader applications of our proposed tensor decomposition
strategy in other areas of media security and forensics.

\bibliographystyle{IEEEtran}
\bibliography{IEEEfull,std_tdsc_submit}

\begin{IEEEbiography}[{\includegraphics[width=1in,height=1.25in,clip,keepaspectratio]{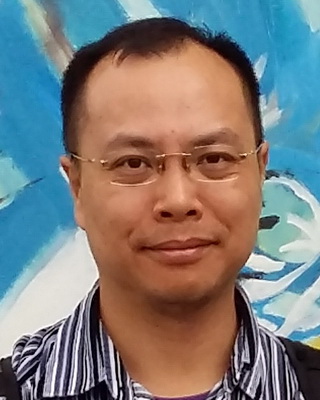}}]{Shunquan Tan (M'10--SM'17)}
  received the B.S. degree in computational mathematics and applied
  software and the Ph.D. degree in computer software and theory from
  Sun Yat-sen University, Guangzhou, China, in 2002 and 2007,
  respectively.

  He was a Visiting Scholar with New Jersey Institute of Technology,
  Newark, NJ, USA, from 2005 to 2006. He is currently an Associate
  Professor with College of Computer Science and Software Engineering,
  Shenzhen University, China, which he joined in 2007. He is the Vice
  Director with the Shenzhen Key Laboratory of Media Security. His
  current research interests include multimedia security, multimedia
  forensics, and machine learning.
\end{IEEEbiography} 

\begin{IEEEbiography}
  [{\includegraphics[width=1in,height=1.25in,clip,keepaspectratio]{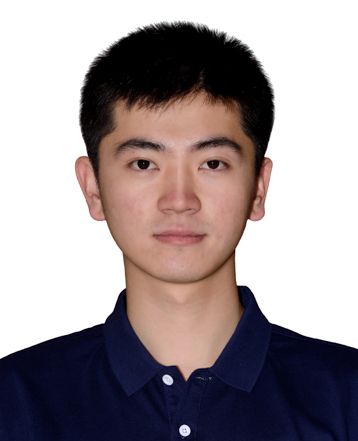}}]{Qiushi Li} 
received the B.S. degree in information and computing science
from Harbin University of Science and Technology, Harbin, China, in 2018.
He is currently pursuing the Ph.D. degree in information and communication
engineering with Shenzhen University, Shenzhen, China. 
His current research interests include multimedia forensics
and machine learning.
\end{IEEEbiography}

\begin{IEEEbiography}[{\includegraphics[width=1in,height=1.25in,clip,keepaspectratio]{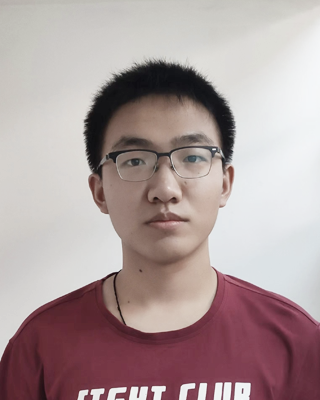}}]{Laiyuan
    Li} is currently a master student in Shenzhen University majoring
  in computer technology. His current research interests include
  multimedia forensics, Image tampering and deep learning.
\end{IEEEbiography}

\begin{IEEEbiography}[{\includegraphics[width=1in,height=1.25in,clip,keepaspectratio]{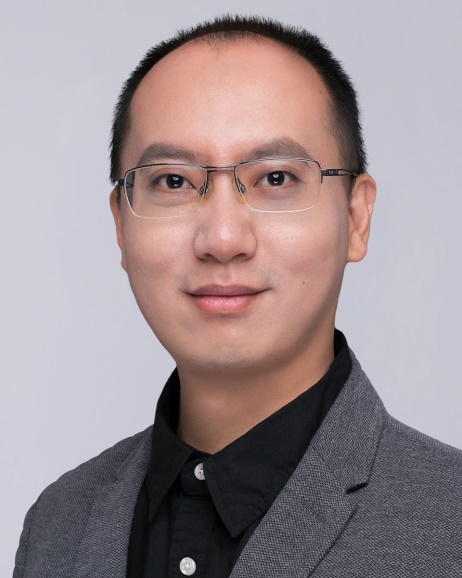}}]{Bin Li (S'07-M'09-SM'17)}
  received the B.E. degree in communication engineering and the Ph.D.
  degree in communication and information system from Sun Yat-sen
  University, Guangzhou, China, in 2004 and 2009, respectively.

  He was a Visiting Scholar with the New Jersey Institute of
  Technology, Newark, NJ, USA, from 2007 to 2008. He is currently a
  Professor with Shenzhen University, Shenzhen, China, where he joined
  in 2009. He is also the Director with the Guangdong Key Lab of
  Intelligent Information Processing and the Director with the
  Shenzhen Key Laboratory of Media Security. He is an Associate Editor
  of the IEEE Transactions on Information Forensics and Security. His
  current research interests include multimedia forensics, image
  processing, and deep machine learning.
\end{IEEEbiography}

\begin{IEEEbiography}[{\includegraphics[width=1in,height=1.25in,clip,keepaspectratio]{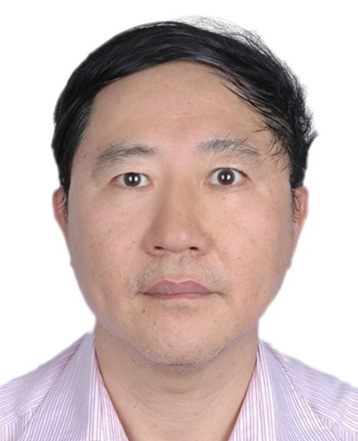}}]{Jiwu Huang (M'98--SM'00--F'16) }
  received the B.S. degree from Xidian University, Xi’an, China, in
  1982, the M.S. degree from Tsinghua University, Beijing, China, in
  1987, and the Ph.D. degree from the Institute of Automation, Chinese
  Academy of Science, Beijing, in 1998. He is currently a Professor
  with the College of Electronics and Information Engineering,
  Shenzhen University, Shenzhen, China. Before joining Shenzhen
  University, he has been with the School of Information Science and
  Technology, Sun Yat-sen University, Guangzhou, China, since
  2000. His current research interests include multimedia forensics
  and security. He is an Associate Editor of the IEEE Transactions on
  Information Forensics and Security.
\end{IEEEbiography}
\vfill
\end{document}